\documentclass[11pt, a4paper, logo]{gdmstyle/googledeepmind}

\pdftrailerid{redacted}

\makeatletter
\renewcommand\bibentry[1]{\nocite{#1}{\frenchspacing\@nameuse{BR@r@#1\@extra@b@citeb}}}
\makeatother

\usepackage{kantlipsum, lipsum}
\usepackage{dsfont}
\usepackage{gdmstyle/gdm-colors}

\usepackage{ifthen}
\newboolean{isanon}
\setboolean{isanon}{false}

\newboolean{isdouble}
\setboolean{isdouble}{false}
\usepackage{xcolor}
\usepackage{xurl}
\usepackage{graphicx}
\usepackage{amsmath}

\usepackage{pifont}%
\usepackage{cleveref}
\usepackage{scalerel}
\usepackage{caption}
\usepackage{wrapfig}
\usepackage{subcaption}
\usepackage{bbding}
\usepackage{placeins}
\usepackage{microtype}
\usepackage{csquotes}
\usepackage{enumitem}
\usepackage{xspace}
\usepackage{booktabs}
\usepackage{mathtools}
\usepackage{amsthm}

\usepackage{tikz}
\usepackage[super]{nth}
\usepackage{multirow}
\usepackage{rotating}
\usepackage{amssymb}
\usepackage[export]{adjustbox}
\usepackage{subcaption}
\usepackage{tcolorbox}

\usetikzlibrary{shapes.geometric, positioning, arrows, shapes.symbols, calc}

\renewcommand*{\backrefalt}[4]{%
    \ifcase #1 \footnotesize{(Not cited.)}%
    \or        \footnotesize{(p.~#2)}%
    \else      \footnotesize{(pp.~#2)}%
    \fi}
\newtheorem{observation}{Observation}

\Crefname{observation}{Observation}{Observations}
\Crefname{hypothesis}{Hypothesis}{Hypotheses}
\newtheorem{remark}{Remark}

\newtheorem*{assumption*}{Assumption}
\Crefname{assumption}{Assumption}{Assumptions}

\newtheorem*{lemma*}{Lemma}
\Crefname{lemma}{Lemma}{Lemmas}

\Crefname{remark}{Remark}{Remarks}
\Crefname{proposition}{Proposition}{Propositions}

\newcommand*\D{\mathcal{D}}

\newcommand*\Loss{\mathcal{L}}

\newcommand*{\versus}{vs.\@\xspace}

\newcommand*{\ie}{i.e.,\@\xspace}

\newcommand*{\wrt}{w.r.t.\@\xspace}

\newcommand*{\aka}{a.k.a.\@\xspace}

\let\originalleft\left
\let\originalright\right
\renewcommand{\left}{\mathopen{}\mathclose\bgroup\originalleft}
\renewcommand{\right}{\aftergroup\egroup\originalright}

\def\eqref#1{eq.~\ref{#1}}

\def\rmI{{\mathbf{I}}}

\DeclareMathAlphabet{\mathsfit}{\encodingdefault}{\sfdefault}{m}{sl}
\SetMathAlphabet{\mathsfit}{bold}{\encodingdefault}{\sfdefault}{bx}{n}

\def\gN{{\mathcal{N}}}

\def\sR{{\mathbb{R}}}

\newcommand{\Esp}{\mathbb{E}}

\newcommand{\acc}{\mathrm{Acc}}
\newcommand{\1}{\mathds{1}}

\newcommand{\KL}{$\mathrm{KL}$\@\xspace}
\newcommand{\robust}{\textcolor{colorredfull}{robust}\@\xspace}
\newcommand{\efficient}{\textcolor{colorbluefull}{efficient}\@\xspace}
\newcommand{\reliable}{\textcolor{coloryellowfull}{reliable}\@\xspace}
\newcommand{\reliably}{\textcolor{coloryellowfull}{reliably}\@\xspace}
\newcommand{\reliability}{\textcolor{coloryellowfull}{reliability}\@\xspace}
\newcommand{\robustness}{\textcolor{colorredfull}{robustness}\@\xspace}
\newcommand{\efficiency}{\textcolor{colorbluefull}{efficiency}\@\xspace}
\newcommand{\efficiently}{\textcolor{colorbluefull}{efficiently}\@\xspace}
\newcommand{\WARM}{\emph{WARM}\@\xspace}
\newcommand{\BAK}{\emph{Baklava}\@\xspace}

\usepackage[sort,numbers,square]{natbib}

\title{\WARM: On the Benefits of Weight Averaged Reward Models}

\correspondingauthor{alexandrerame@google.com}


\author{Alexandre~Ramé}
\author{Nino~Vieillard}
\author{Léonard~Hussenot}
\author{Robert~Dadashi}
\author{Geoffrey~Cideron}
\author{Olivier~Bachem}
\author{Johan~Ferret}
\affil{Google DeepMind}

\begin{abstract}
	Aligning large language models (LLMs) with human preferences through reinforcement learning (RLHF) can lead to reward hacking, where LLMs exploit failures in the reward model (RM) to achieve seemingly high rewards without meeting the underlying objectives.
We identify two primary challenges when designing RMs to mitigate reward hacking: distribution shifts during the RL process and inconsistencies in human preferences.
As a solution, we propose Weight Averaged Reward Models (\WARM), first fine-tuning multiple RMs, then averaging them in the weight space.
This strategy follows the observation that fine-tuned weights remain linearly mode connected when sharing the same pre-training.
By averaging weights, \WARM improves \efficiency compared to the traditional ensembling of predictions, while improving \reliability under distribution shifts and \robustness to preference inconsistencies.
Our experiments on summarization tasks, using {best-of-$N$} and RL methods, shows that \WARM improves the overall quality and alignment of LLM predictions; for example, a policy RL fine-tuned with \WARM has a 79.4\% win rate against a policy RL fine-tuned with a single RM.

\end{abstract}

\begin{document}

\maketitle

\section{Introduction}

\textbf{Reward modeling.}
Conversational assistants such as Gemini \cite{gemini2023} or GPT-4 \cite{openai2023gpt4} have revolutionized the AI community and beyond.
These LLMs are capable of completing novel and intricate tasks, including mathematics, coding, and tool use \cite{bubeck2023sparks}.
These advancements are underpinned by a systematic three stage training procedure: pre-training by next token prediction \cite{radford2018gen,devlin2018bert,NEURIPS2020_1457c0d6}, supervised fine-tuning (SFT) to learn to follow instructions \cite{wei2022finetuned,wang-etal-2022-super,alpaca2023}, and ultimately, reinforcement learning (RL) to maximize a reward encapsulating the desired behaviors \cite{roit2023factually}.
However, defining such rewards for real-world tasks is non-trivial~\cite{mckinney2023fragility}.
In reinforcement learning from human feedback (RLHF) \cite{christiano2017deep,ziegler2019fine,stiennon2020learning,wu2021recursively}, rewards are reward models (RMs), trained on binary preference datasets to emulate human judgment.
The enhancement of LLM capabilities from RL is strongly tied to the quality of the RMs \cite{touvron2023llama2}.

\begin{figure*}[t!]
	\begin{center}
		\begin{subfigure}[b]{0.66\textwidth}
			\includegraphics[width=\textwidth]{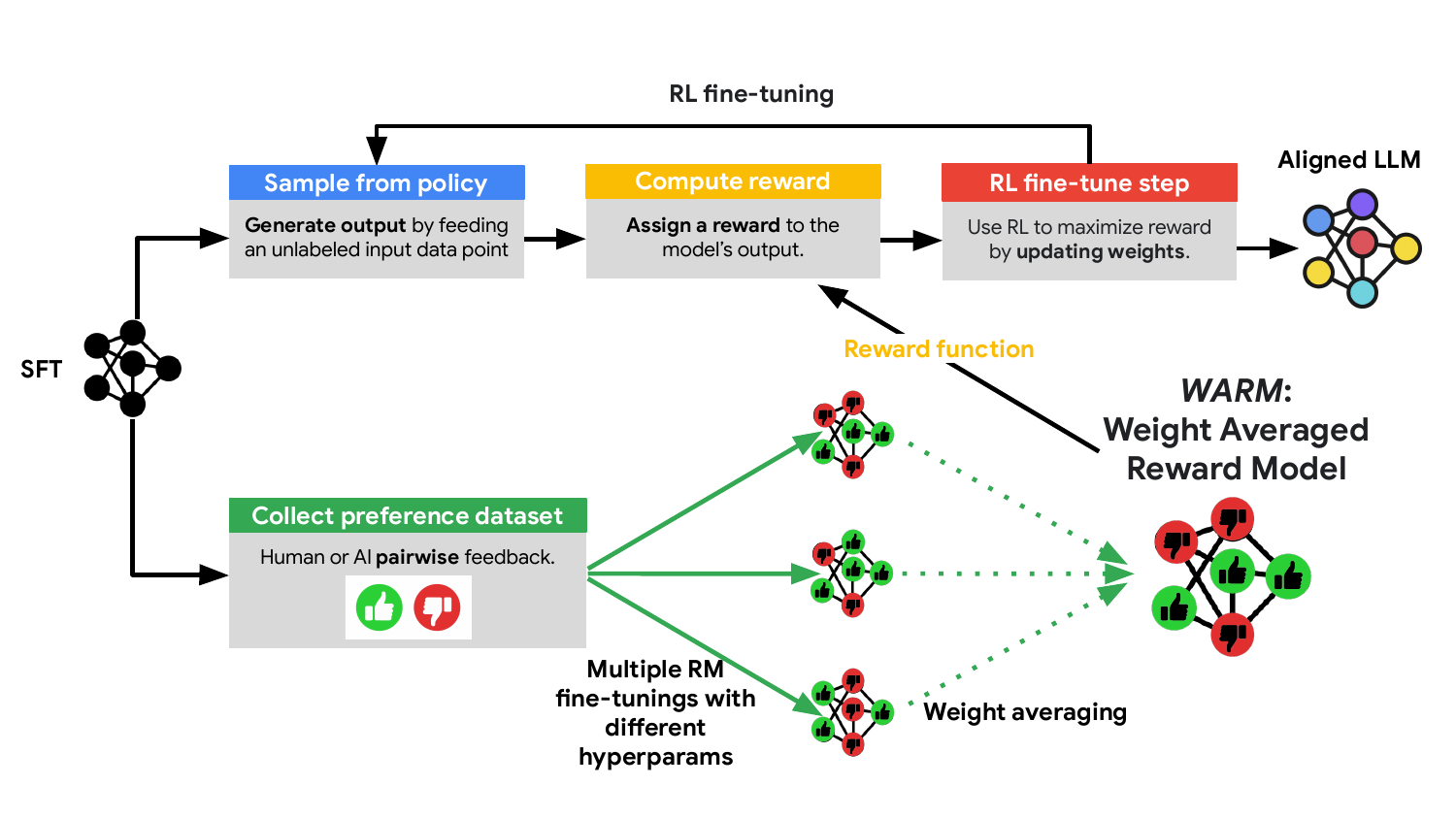}
			\caption{\WARM procedure with $M=3$.}
			\label{fig:main:warm}%
		\end{subfigure}%
		\hfill%
		\begin{subfigure}[b]{0.33\textwidth}%
			\includegraphics[width=\textwidth]{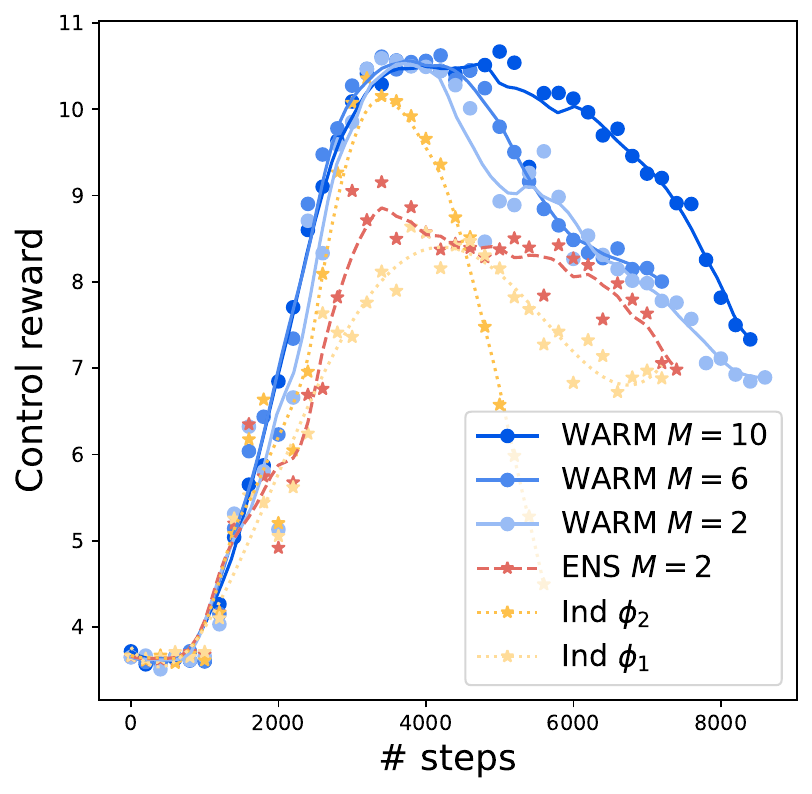}
			\caption{\WARM mitigates reward hacking.}%
			\label{fig:main:hacking}%
		\end{subfigure}
	\end{center}%
	\caption{\Cref{fig:main:warm} illustrates the alignment process with \WARM. From a SFT-ed LLM, we apply RL fine-tuning to optimize a proxy reward model (RM), in line with RLHF \cite{christiano2017deep}. The innovation of \WARM lies in the design of the proxy RM, which is the weight average (WA) of $M$ individual RMs, each fine-tuned from a shared pre-trained LLM on the same preference dataset, but with slight differences such as diverse hyperparameters.
	This WA approach is \efficient, while enhancing the \reliability under distribution shifts and \robustness under inconsistent preferences.
	\Cref{fig:main:hacking} showcases the impact during RL alignment. The control reward (detailed in \Cref{sec:expe}) initially increases but eventually deteriorates, a phenomenon called reward hacking \cite{amodei2016concrete}. However, when \WARM serves as the proxy RM, increasing $M$ (the number of averaged RMs) significantly improves absolute results while delaying the collapse, as indicated by the control rewards maintaining higher values for longer during training. Same plot with \KL as the $x$-axis in \Cref{fig:hackingkl} and with label corruption in \Cref{fig:controlvsstep_corruptalp}.}%
	\label{fig:main}%
\end{figure*}

\textbf{Reward hacking.}
Particularly insidious in RLHF \cite{gao2022scaling,casper2023open} is the \textit{reward hacking} issue \cite{amodei2016concrete,faultyreward2016,askell2021general,skalse2022defining} (\aka reward overoptimization), arising from \textit{reward misspecification} \cite{pan2022the,lambert2023alignment} between the proxy RM and actual human preferences.
While optimizing for the RM initially provides improvements, in later stages the policy (\ie the LLM being trained) usually learns to exploit loopholes in the RM and achieves high rewards without truly fulfilling the intended objectives, as illustrated in \Cref{fig:main:hacking}.
This reward hacking phenomenon poses numerous issues.
First, it degrades performances, manifesting as linguistically flawed \cite{lewis2017deal} or unnecessarily verbose \cite{singhal2023long} outputs, which do not reflect true human preferences.
Second, it complicates checkpoint selection due to the unreliability of the proxy RM, echoing Goodhart's Law \cite{Strathern1997}: \enquote{when a measure becomes a target, it ceases to be a good measure}.
Third, it can engender sycophancy \cite{perez2022discovering,sharma2023towards} or amplify social biases, reflecting the limited and skewed demographics of feedback providers \cite{santurkar2023whose,hartmann2023political}.
Lastly and most critically, misalignment \cite{taylor2016alignment,ngo2022alignment} due to reward hacking can escalate into safety risks~\cite{amodei2016concrete,hendrycks2022x,hendrycks2023natural}, in particular given the rapid integration of LLMs in everyday life and critical decision-making.
Such concerns underscore the need to mitigate reward hacking to ensure the beneficial and safe deployment of LLMs.

\textbf{Challenges.}
Two primary challenges underlie reward hacking.
The first major issue are \textit{the distribution shifts} encountered by the RM \cite{zhuang2020consequences,shin2023benchmarks}.
Indeed, the generations from the policy might deviate substantially from those in the offline preference dataset, posing an out-of-distribution (OOD) challenge.
Moreover, those distribution shifts are accentuated by the policy drift during the RL procedure: the policy moves away from its SFT initialization, continually altering the distribution of predictions the RM needs to interpret \reliably.
Second, \textit{preferences are inconsistent}: the binary labels in the preference dataset are noisy.
Indeed, human labelers often rely on simpler criteria (length, bullet points, politeness) over more nuanced indicators. Moreover, errors can be exacerbated for complex tasks requiring specific expertise \cite{bowman2022measuring}, and because of the multi-objective nature of alignment \cite{rame2023rewarded} requiring handling the heterogeneity of human opinions. Overall, this results in a low inter-labeler agreement (72.6\% for InstructGPT \cite{ouyang2022training}), altering the \robustness of the RM.

\textbf{Goal and ensembling baseline.}
Designing good RMs must meet a tripartite objective: guiding RL \efficiently, \reliably scoring generations despite the distribution shifts, and providing \robust signals amidst label noise.
To address these challenges, the seminal work on RLHF from Christiano \textit{et al.} \cite{christiano2017deep} and more recent works \cite{ensemble2023reward,coste2023reward} leveraged \textit{prediction ensembling} (ENS) \cite{Lakshminarayanan2017}, averaging the rewards from multiple RMs. ENS improves the \reliability of the reward and mitigates hacking risks \cite{ensemble2023reward,coste2023reward}. Yet, ENS suffers from memory and inference overhead causing \efficiency challenges; we will also show that ENS fails to improve \robustness to label noise in the preference datasets.

\textbf{\WARM.}
In this paper, we propose weight averaged reward models (\WARM), a simple, \efficient and scalable strategy to obtain a \reliable and \robust RM by combining multiple RMs.
Starting from a shared pre-trained LLM, we launch multiple RM fine-tunings: in practice, the different runs have different hyperparameters (as in grid search), and see the preference data in different orders, thus leading to diverse RMs.
A key contribution is how the different RMs are merged: by \textit{linear interpolation in the weight space}.
This follows the findings from the linear mode connectivity (LMC)~\cite{Frankle2020,Neyshabur2020} and weight averaging (WA) literature \cite{Wortsman2022ModelSA,rame2022diwa,rame2022recycling}: under shared pre-training, the different weights can be linearly interpolated despite the non-linearities in the architecture.

\textbf{On the benefits of \WARM.}
Firstly, \WARM stands out for its \efficiency and practicality. By requiring a single model at inference time, it provides a scalable approximation to the traditional, costlier ensembling of predictions, without its memory and inference burdens.
Secondly, \WARM improves \reliability by inheriting from the generalization abilities of WA under distribution shifts, a quality well-documented in the OOD literature for supervised learning \cite{cha2021wad,rame2022diwa,rame2022recycling}.
Lastly, \WARM improves \robustness to label corruption.
We show that WA selects the invariant predictive mechanisms \cite{muandet2013domain,arjovsky2019invariant} across different runs~\cite{zaman2023fuse,lin2023spurious}, thus naturally diminishing the memorization of corrupted samples, occurring in each run in different ways. In contrast, ENS simply memorizes the corrupted samples. We also explain why reducing memorization when modeling noisy preferences enhances stability in the RL process. These multifaceted benefits of \WARM are further explored in \Cref{sec:benefits}.

We summarize our contributions as follows.
\begin{enumerate}
    \item \textit{Innovation in reward modeling}. We introduce \WARM, the first instance of weight averaging for reward modeling. This novel strategy \efficiently mitigates reward hacking, improves \reliability under distribution shifts and \robustness to label corruption.
    \item \textit{Theoretical and empirical insights into weight averaging}. We validate linear mode connectivity for reward models trained on binary preference datasets. Moreover, we reveal a key difference between weight and prediction averaging, that appears clearly under label corruption; weight averaging only maintains the invariant predictive mechanisms across runs, thereby diminishing memorization and enhancing the focus on generalizable features.
\end{enumerate}
Our experiments on summarization tasks in \Cref{sec:expe} confirm that \WARM improves performance without any memory or inference overhead, either when used as the reward selector in best-of-$N$, or as the proxy RM in RL.
\WARM mitigates reward hacking, and thus provides better downstream policies; specifically, it leads to a win rate of 79.4\% (according to the preference oracle metric) against a policy trained with a standard RM.

\section{Context and challenges}
\label{sec:context}
\subsection{Context}
\textbf{LLMs.}
We consider an LLM $f_{\theta}$ of a fixed non-linear architecture parameterized by $\theta$, usually a Transformer with attention layers \cite{vaswani2017transformer}. It defines a policy by mapping prompt inputs $x$ to $f_{\theta}(x)$.
Following the foundation model paradigm \cite{bommasani2021opportunities} and the success of transfer learning \cite{oquab2014learning}, the weights $\theta$ are first pre-trained \cite{radford2018gen} on the vast amount of web data into $\theta^{pt}$, before supervised fine-tuning (SFT)~\cite{wei2022finetuned} to learn to follow instructions into $\theta^{sft}$. However, the high cost and limited scope of instruction data (\ie prompts and responses) can create a misalignment \cite{amodei2016concrete,taylor2016alignment,ngo2022alignment} between the LLM and its intended application. Reinforcement learning (RL) as a third step in the training process of LLMs was shown to help alignment of LLMs with the intended usage \cite{ouyang2022training}.

\textbf{RMs.}
A notable aspect of RL is the absence of supervised samples to be imitated by the policy; instead, the focus shifts to maximizing the reward of generated samples, that should measure their quality.
The challenge is that the oracle reward, perfectly encapsulating the desired behaviors, is not given by the environment.
The key innovation from RLHF \cite{christiano2017deep} is that this reward is the output of a reward model (RM), trained in a supervised way to predict and thus reflect human preferences. Specifically, an RM is an LLM $r_{\phi}$ parameterized by $\phi$, predicting a single scalar as the reward $r_{\phi}(x, y)$ for a prompt $x$ and generation $y$. The weights $\phi$ are usually initialized from $\left(\theta^{sft}, \omega\right)$, where the final linear layer $\omega$ is added on top of the extracted features from the SFT model $\theta^{sft}$.
Then $\phi$ is trained on a preference dataset $\D_{train}=\{x_d, y_d^{+},y_d^{-}\}_{d=1}^{D}$ where the generation $y_d^{+}$ has been preferred over $y_d^{-}$ to continue $x_d$.
Usually human labelers evaluate those generations, but recent works on RLAIF \cite{bai2022constitutional,lee2023rlaif} showed that similar performances can be obtained by prompting an LLM for AI feedback.
Following the Bradley-Terry \cite{bradley1952rank} assumption about the distribution of preferences, and by framing the problem as binary classification, the maximum likelihood principle motivates learning $\phi$ by minimizing the following negative log-likelihood loss (where $\sigma$ is the logistic function):
\begin{equation}
	\Loss_R\left(r_{\phi}, \D_{train}\right)=-\Esp_{\left(x, y^{+}, y^{-}\right) \in \D_{train}}\left[\log \sigma\left(r_{\phi}\left(x, y^{+}\right)-r_{\phi}\left(x, y^{-}\right)\right)\right].%
	\label{eq:rm}%
\end{equation}.

\textbf{Reward inference.}
With this RM, the literature suggests applying any kind of RL algorithm (usually REINFORCE \cite{williams1992simple} or PPO \cite{schulman2017proximal}) to fine-tuned $\theta^{sft}$ into $\theta^{rl}$, as analyzed in \Cref{sec:expe:rl}.
A training-free alternative is best-of-$N$ (BoN) sampling, analyzed in \Cref{sec:expe:bon}, which returns the generation that has the highest reward among $N$ generations from $\theta^{sft}$.
Both methods aim to align the policy with human preferences. Yet, the \textit{reward misspecification} \cite{pan2022the} between the proxy RM and the true human preferences can lead to \textit{reward hacking} \cite{amodei2016concrete,faultyreward2016,askell2021general,skalse2022defining}, where the policy exploits loopholes in the proxy RM to artificially increase the score without matching human preferences.

\subsection{Challenges in reward modeling}
When handling rich inputs such as text, or when assessing complex behaviours, designing rewards aligned with human preferences is a complex challenge for two main reasons, described below.

\textbf{Distribution shifts.}
The primary challenge is the distribution shifts resulting from the offline nature of preference data.
Indeed, the generations in the preference dataset and those from the policy $\theta^{sft}$ do not necessarily follow the same distributions, and the shifts can become even more pronounced due to model drift during RL.
The OOD generalization literature has extensively analyzed the repercussions of these shifts.
Firstly, they often lead to a reduction in performance \cite{gulrajani2021in,pmlr-v139-koh21a}. RMs (of limited capacity) trained on narrow data distributions may rely on spurious correlations \cite{arjovsky2019invariant} or a limited number of features \cite{pezeshki2020radient}, thus failing when encountering OOD examples \cite{laakom2021learning,nayman2022diverse}.
Secondly, they complicate the selection of RMs, as ID validation metrics may poorly correlate with real-world OOD performances \cite{d2020underspecification,teney2022id} and the ability to guide the RL \cite{ensemble2023reward}.
Lastly, RMs can become poorly calibrated \cite{guo2017calibration} in OOD scenarios \cite{Ovadia2019,wald2021n}, and predict more extreme values as rewards. Such miscalibration exacerbates the problem in a negative feedback loop, further intensifying model drift and distribution shifts.
In conclusion, limited data coverage during reward modeling reduces the \reliability of the RM and facilitates reward hacking \cite{zhuang2020consequences} in regions where the RM is badly specified.

\textbf{Inconsistent preferences.}
The second major challenge is the label noise in preference datasets.
Human labelers, often grappling with fatigue, misunderstandings \cite{simon1990bounded,shah2019feasibility} and imperfect incentives \cite{kaufmannchallenges}, might default to simpler criteria such as length, bullet points, or politeness rather than more causal indicators. This tendency is exacerbated for complex tasks \cite{bowman2022measuring} or when considering multiple objectives, ranging from harmlessness \cite{ganguli2022red} to engagement \cite{irvine2023rewarding} and representing the heterogeneity of human opinions.
Consequently, these factors lead to low inter-rater agreement, where human data appears as an imperfect representation of the underlying ground truth \cite{condorcet1785,pitis2023failure}.
To mitigate these issues, there has been a shift towards AI-generated preferences \cite{bai2022constitutional,lee2023rlaif}, which, while reducing human labor costs, introduces its own set of noise and failure cases, such as sensitivity to prompting strategies \cite{sclar2023quantifying,mizrahi2023state}.
These layers of noise and inconsistency challenge the \robustness of the RM, and its ability to provide stable signals.

With this in mind, a good RM should ideally satisfy the three following properties.
\begin{tcolorbox}[colback=colorblue,
		colframe=black,
		arc=4pt,
		boxsep=0.5pt,
	]
	\textbf{\textit{Property} 1: \textit{\efficiency}.} The RM should incur no memory or inference overhead. Then the policy can be optimized efficiently.%
\end{tcolorbox}

\begin{tcolorbox}[colback=coloryellow,
		colframe=black,
		arc=4pt,
		boxsep=0.5pt,
	]
	\textbf{\textit{Property} 2: \textit{\reliability}.} The RM should reliably reward predictions despite the distribution shifts. Then the policy can explore away from its initialization while relying on the RM.
\end{tcolorbox}

\begin{tcolorbox}[colback=colorred,
		colframe=black,
		arc=4pt,
		boxsep=0.5pt,
	]
	\textbf{\textit{Property} 3: \textit{\robustness}.} The RM should be robust to the label inconsistencies in binary preferences. Then the policy can learn from robust signals given by the RM.
\end{tcolorbox}

\subsection{Existing approaches}
To tackle those issues, previous works have explored a few research directions, further detailed in our related work from \Cref{sec:related:rm}.
During RL, the standard strategy is to encourage the policy to remain close to its SFT initialization with Kullback-Leibler (\KL) regularization \cite{jaques2017sequence,geist2019theory}; \KL reduces model drift \cite{lazaridou2020multi,lu2020countering} but can cause underfitting and adds an extra hyperparameter (the regularization strength $\alpha$).
Collecting, labelling and then training on new data (reflecting the evolving policy) can improve the \reliability of the RM \cite{touvron2023llama2}. Yet it poses significant \efficiency challenges due to the continuous requirement for human annotation and computational resources. In contrast, \textit{active learning} strategies \cite{reddy2020learning,10.5555/3237383.3238074} proactively enrich the preference dataset by seeking out a diverse set of generations and potential failure cases.
Concurrent work \cite{secret2023rm} suggests applying label smoothing and flipping.
Finally, and most similar to \WARM, \textit{prediction ensembling} (ENS) \cite{Lakshminarayanan2017} strategies average the logits from $M$ RMs.
From a bias-variance perspective \cite{kohavi1996bias}, ENS reduces the variance term when members are sufficiently diverse \cite{ueda1996generalization}, and thus favors \reliability under distribution shifts where variance is the key issue \cite{rame2022diwa}.
From a RL perspective, ENS was shown to mitigate hacking risks \cite{christiano2017deep,coste2023reward,ensemble2023reward}.
Despite its advantages, ENS faces \efficiency challenges; the memory and inference costs grow linearly with $M$, making ENS incompatible with the scaling trend in RMs, where larger architectures consistently perform better \cite{kundu2023specific}.
Moreover, we will also show in \Cref{sec:expe:labelnoise} that ENS fails to improve \robustness to preference inconsistencies.
\section{\WARM}
\label{sec:model}
\subsection{Weight averaging of reward models}%
Facing those challenges in reward modeling and the limitations from existing approaches, we propose Weight Averaged Reward Models (\WARM). \WARM is a simple and \efficient strategy that combines multiple models without the memory and inference overheads of prediction ensembling, enhancing reward \reliability (under distribution shifts) and \robustness (amidst noisy preference dataset).
\WARM is illustrated in \Cref{fig:main:warm} and described below.
\begin{enumerate}
	\item \textit{Shared pre-trained initialization}. For a given pre-trained LLM, each RM is initialized from $\left(\theta^{sft}, \omega\right)$ combining SFT weights and a linear probed \cite{kumar2022finetuning} classifier.
	\item \textit{Diverse fine-tunings}. We run $M$ RM fine-tunings, optimizing \Cref{eq:rm} with diverse hyperparameters (as in a grid search), yielding $M$ weights $\{\phi_{i}\}_{i=1}^{M}$.
	\item \textit{Weight averaging}. We average those $M$ weights together to form $\phi^{\text{WARM}}=\frac{1}{M} \sum_{i=1}^{M} \phi_{i}$.
\end{enumerate}%
Then $r_{\phi^{\text{WARM}}}$ serves as the proxy RM to guide the RL procedure, as \efficiently as an individual RM, but with the enhanced \reliability and \robustness provided by the WA strategy, that leverages the strengths and mitigates the weaknesses of the individual RMs.%
\FloatBarrier
\subsection{Linear mode connectivity}
Compared to ENS, the main difference lies in how \WARM combines the different RMs: we do so through \textit{linear interpolation in the weight space}.
It relies on the linear mode connectivity (LMC)~\cite{Frankle2020,Neyshabur2020} property across fine-tuned weights, \ie the fact that the accuracy of the interpolated model is at least as good as the interpolation of the individual accuracies.
Precisely, by defining the pairwise accuracy of an RM $r_{\phi}$ \wrt a dataset $\D$ as $\acc\left(r_{\phi}, \D\right)=\Esp_{\left(x, y^{+}, y^{-}\right) \in \D}\left[\1_{r_{\phi}(x, y^{+})\geq r_{\phi}(x, y^{-})}\right]$, the following \Cref{obs:lmc} underpins the success of \WARM.
\begin{observation}[LMC]
	Given two fine-tuned weights $\phi_1$ and $\phi_2$ with a shared pre-training and a test dataset $\D_{test}$, then for all $\lambda \in [0, 1]$,%
	\begin{equation}
		\acc\left(r_{\left(1 - \lambda\right) \cdot \phi_1 + \lambda \cdot \phi_2}, \D_{test}\right) \geq \left(1 - \lambda\right) \times \acc\left(r_{\phi_1}, \D_{test}\right) + \lambda \times \acc\left(r_{\phi_2}, \D_{test}\right).%
	\end{equation}%
	\label{obs:lmc}%
\end{observation}%
We empirically validate this LMC in \Cref{fig:lmc_wavsens}, by evaluating interpolated RMs on OOD test samples.
This follows similar observations for multi-class classification in the context of computer vision~\cite{Frankle2020,Neyshabur2020}, which led to a plethora of weight averaging (WA) works such as the model soups \cite{Wortsman2022ModelSA,rame2022diwa,rame2022recycling} variants (detailed in our related work in~\Cref{sec:related:wa}).
\begin{remark}[Importance of pre-training and linear probing]%
	The efficacy of WA can be surprising given the non-linearities \cite{vaswani2017transformer} and permutation symmetries \cite{ainsworth2022gitrebasin} in deep neural network architectures.
	WA is actually possible only because of the shared pre-training which constrains the divergence during fine-tunings \cite{Neyshabur2020}, such as the weights remain in convex regions of the loss valley \cite{gueta2023knowledge}.
	In contrast, the LMC does not hold when training weights from scratch \cite{Neyshabur2020}, even if the random initialization is shared.
    For these reasons and to facilitate the LMC, we follow \cite{rame2022diwa,rame2022recycling} and use linear probing to initialize the classifier $\omega$; compared to random initialization, such linear probing prevents feature distortion~\cite{kumar2022finetuning}.%
\end{remark}%
\subsection{Sources of diversity}
On one hand, \WARM requires shared pre-training so that the fine-tuned weights remain linearly connected.
On the other hand, weights must not be identical: actually, the diversity across those fine-tuned weights significantly contributes to the accuracy gains observed in WA \cite{rame2022diwa}.
Overall, an effective \WARM requires a delicate trade-off between ensuring LMC and diversity across weights.

In practice, we use the following sources of diversity \cite{gontijolopes2022no}, leading the RM fine-tunings to \textit{diverse yet linearly connected} models.
First, the different fine-tunings see the data samples in \textit{different orders}.
Second, we sample slightly \textit{different hyperparameters}, notably different learning rates and dropout probabilities, as detailed in \Cref{app:details:rm}.
Third, we investigate a new source of \textit{diversity in initialization} named \BAK, illustrated in \Cref{fig:baklava}. Specifically, we initialize the RMs' featurizers from different checkpoints $\{\theta_i^{sft}\}_{i=1}^M$ collected along a given SFT trajectory. \BAK relaxes the shared initialization constraint from model soups \cite{Wortsman2022ModelSA} to simply sharing the same pre-training: \BAK is actually an \efficient alternative to model ratatouille \cite{rame2022recycling} but without the need of multiple auxiliary tasks.
Overall, \BAK increases diversity compared to only initializing from the last SFT checkpoint, while adhering to the shared pre-training requisite for LMC, without incurring any overhead.%
\begin{figure*}[t!]
	\begin{center}
		\resizebox{0.6\textwidth}{!}{\begin{tikzpicture}
    \newcommand{\YA}{2}
    \newcommand{\YB}{4}
    \newcommand{\YC}{6}
    \newcommand{\YD}{8}    
    \newcommand{\YE}{10}  
    \newcommand{\YF}{10}  
    \node at (1, 0) (a) {SFT};
    \node at (1, -0.8) (a) {Reward};
    \node at (1, -1.2) (a) {modelings};    
    \node at (1, -1.8) (a) {Weight};
    \node at (1, -2.2) (a) {averaging};    
    \node at (\YA, 0.4) (a) {$\theta^{pt}$};
    \node at (\YB, 0.4) (a) {$\theta_1^{sft}$};
    \node at (\YC, 0.4) (a) {$\theta_2^{sft}$};
    \node at (\YD, 0.4) (a) {$\theta_M^{sft}$};    

    \node[draw, fill=colorbluefull!10] at (\YA, 0) (a0) {};
    \node[draw, fill=colorbluefull!20] at (\YB, 0) (a1) {};
    \node[draw, fill=colorbluefull!30] at (\YC, 0) (a2) {};
    \node[draw, fill=colorbluefull!40] at (\YD, 0) (a3) {};

    \node at (5.5, -1) (a) {$\phi_1$};
    \node at (7.5, -1) (a) {$\phi_2$};
    \node at (9.5, -1) (a) {$\phi_M$};
    \node at (7.7, -2) (a) {$\phi^{\text{WARM}}=\frac{1}{M}\sum_{i=1}^M \phi_i$};    
    \node[draw, fill=coloryellowfull!20] at (\YC, -1) (b1) {};
    \node[draw, fill=coloryellowfull!30] at (\YD, -1) (b2) {};
    \node[draw, fill=coloryellowfull!40] at (\YE, -1) (b3) {};
    \node[draw, fill=colorredfull!50] at (\YF, -2) (c4) {};

    \draw[->, dashed] (a0) -- (a3);
    \draw[->, very thick] (a1) -- (b1);
    \draw[->, very thick] (a2) -- (b2);
    \draw[->, very thick] (a3) -- (b3);
    \draw[->, dotted] (b1) -- (c4);
    \draw[->, dotted] (b2) -- (c4);
    \draw[->, dotted] (b3) -- (c4);
  \end{tikzpicture}
  }%
	\end{center}%
	\caption{\textbf{\BAK diversity procedure}.
	Starting from a pre-trained LLM $\theta^{pt}$, we consider different checkpoints $\{\theta_{i}^{sft}\}_{i=1}^{M}$ along a single SFT run (dashed arrow \protect\tikz[baseline]\protect\draw[->, dashed](0ex,0.8ex) -- (3ex,0.8ex);) collected at different number of SFT training steps.
	Those checkpoints serve as initializations for $M$ RM fine-tunings on the preference dataset (thick solid arrows \protect\tikz[baseline]\protect\draw[->, very thick](0ex,0.8ex) -- (3ex,0.8ex);) to learn the $\{\phi_{i}\}_{i=1}^{M}$.
	Finally, those RMs are weight averaged (dotted arrows \protect\tikz[baseline]\protect\draw[->, dotted](0ex,0.8ex) -- (3ex,0.8ex);) into the final model $\phi^{\text{WARM}}$.
	Following the culinary analogy from model soups \cite{Wortsman2022ModelSA} and model ratatouille \cite{rame2022recycling}, we named this method \href{https://en.wikipedia.org/wiki/Baklava}{\BAK} because of its diamond geometric shape.}%
	\label{fig:baklava}%
\end{figure*}
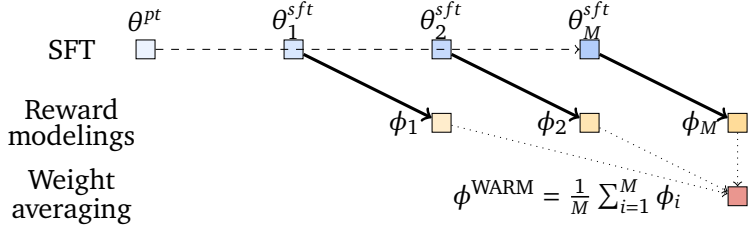
\begin{remark}[Moving average]%
\textcolor{cadetgrey}{Following stochastic weight average \cite{izmailov2018} or moving average \cite{arpit2021ensemble}, we also tried to average checkpoints collected along a single RM fine-tuning.
Though interesting because less costly for training, the lower results in \Cref{fig:lmc_ma8k10k} suggest that the accuracy-diversity trade-off was not favorable: incorporating early checkpoints would compromise individual accuracies, and considering only later checkpoints would not bring the necessary diversity. As a result, we opted to use in \WARM only the last checkpoint from each RM fine-tuning.}
\end{remark}%

\section{On the benefits of \WARM}
\label{sec:benefits}
We now explore the properties and benefits from the \WARM strategy, previously described in \Cref{sec:model}.
We ground our analysis on the empirical comparison between WA and ENS for reward modeling, and a novel general theoretical comparison in \Cref{sec:benefits:theory}.

\textbf{Experimental setup.} We leverage the TL;DR summarization benchmark \cite{volske2017tl}, a standard in reward modeling for LLMs, that we briefly describe below and further detail in \Cref{app:details}.
The goal of the RMs is to score summaries such as they are ranked properly.
In training, we use the dataset $\D_{train}$ from Stiennon \textit{et al.} \cite{stiennon2020learning} where the candidate summaries are generated by GPT-3 \cite{NEURIPS2020_1457c0d6} variants.
To obtain the labels, we follow the RLAIF procedure from \cite{lee2023rlaif}, where a PaLM-L \cite{anil2023palm} is prompted with chain-of-thought \cite{wei2022chain} to generate feedback mimicking human preferences. This strategy performs similarly to human labelers with similar inter-agreement, and will be useful in \Cref{sec:expe} as an oracle metric.
The RMs are PaLM-XXS models, pre-trained and SFT-ed on the preferred summaries from $\D_{train}$, on which we plug a linear probed \cite{kumar2022finetuning} classification layer.
We train the RMs for 10k steps on $\D_{train}$, with hyperparameters and procedure detailed in \Cref{app:details:rm}.
We report accuracies of those RMs on a novel out-of-distribution (OOD) test dataset $\D_{ood}$ with 92k pairwise comparisons where the summaries are generated by multiple PaLM-XS policies with high temperature, some of which are pre-trained only, others SFT-ed and others RLHF-ed.
\subsection{\nth{1} order analysis: weight averaging for \reliable and more \efficient ensembling}
Previous works \cite{izmailov2018,Wortsman2022ModelSA,rame2022diwa} have argued that the best way to understand WA is as an efficient approximation of ENS, as clarified in \Cref{obs:wiens}.%
\begin{observation}[WA and ENS: \nth{1} order analysis]
Weight averaging and prediction ensembling perform similarly: \ie for all $\lambda \in [0, 1]$ and a test dataset $\D_{test}$,%
	\begin{equation}
		\acc\left(r_{\left(1 - \lambda\right) \cdot \phi_1 + \lambda \cdot \phi_2}, \D_{test}\right) \approx \acc\left(\left(1-\lambda\right) \times r_{\phi_1} + \lambda \times r_{\phi_2}, \D_{test}\right).%
	\end{equation}%
	\label{obs:wiens}%
\end{observation}%
\begin{figure*}[t!]
	\begin{center}
		\begin{subfigure}[b]{0.24\textwidth}
			\includegraphics[width=1.0\textwidth]{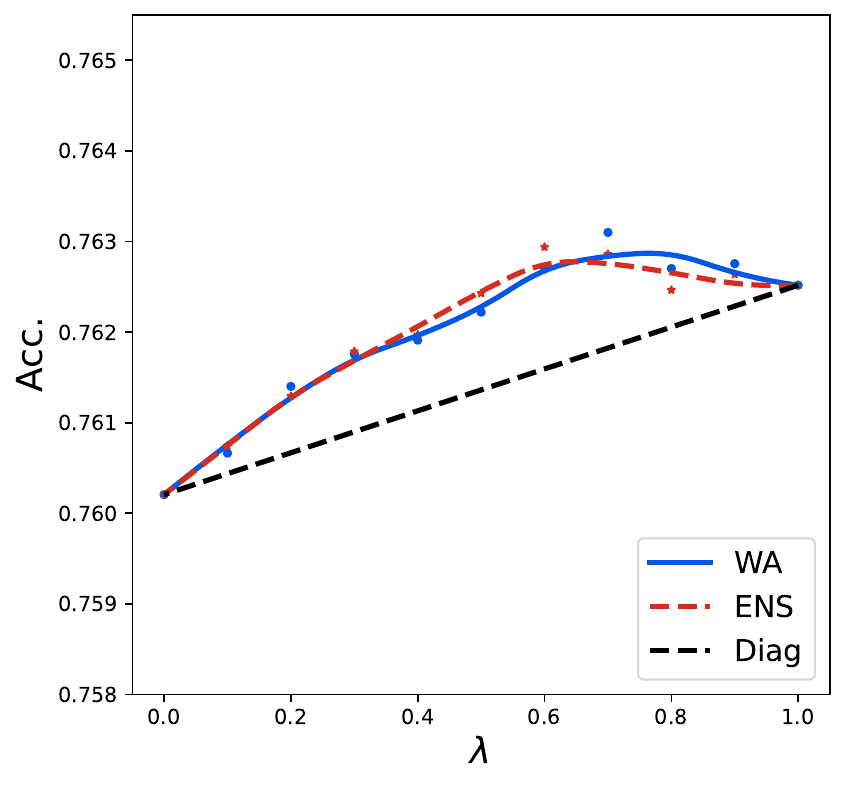}
			\caption{1 RM fine-tuning at 2 different training steps.}
			\label{fig:lmc_ma8k10k}%
		\end{subfigure}%
		\hfill		
		\begin{subfigure}[b]{0.24\textwidth}
			\includegraphics[width=1.0\textwidth]{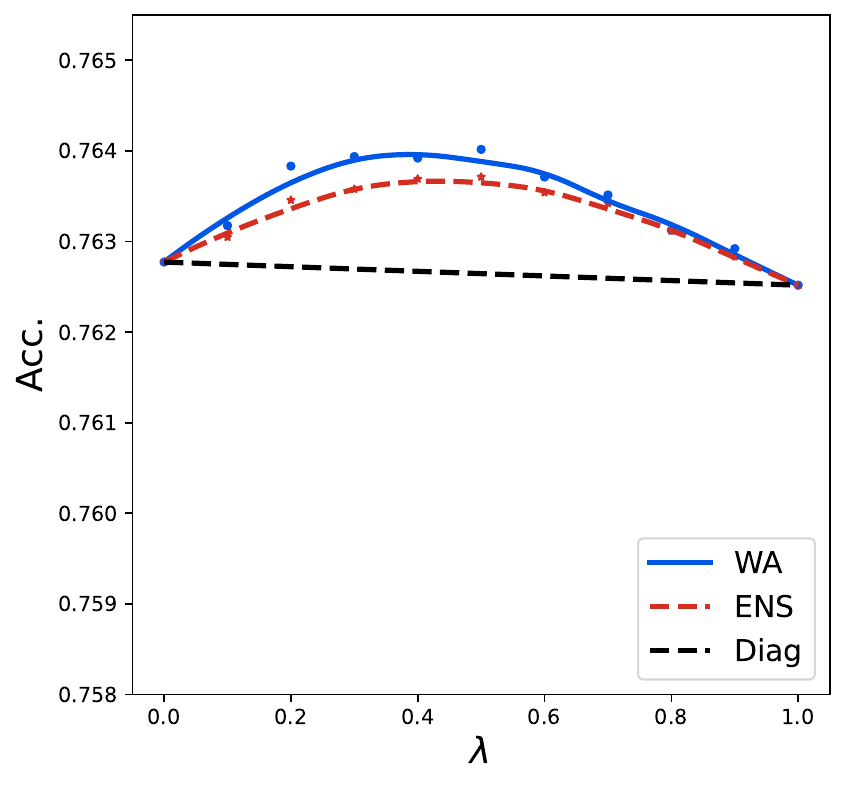}
			\caption{2 RM fine-tunings with shared config.}
			\label{fig:lmc_same}%
		\end{subfigure}%
		\hfill		
		\begin{subfigure}[b]{0.24\textwidth}
			\includegraphics[width=1.0\textwidth]{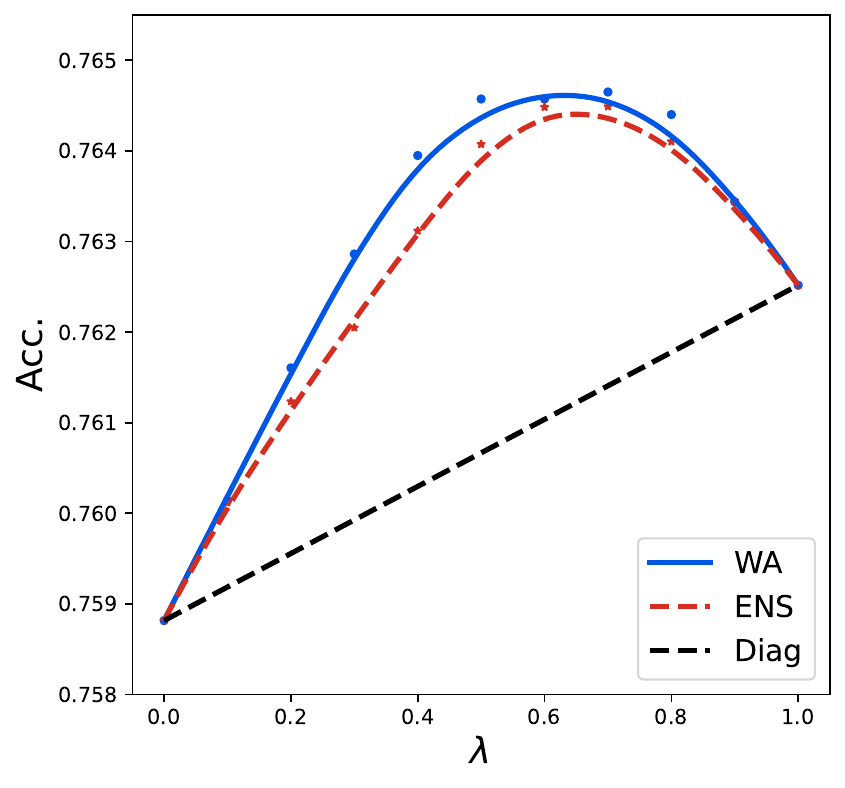}
			\caption{2 RM fine-tunings with different learning rates.}
			\label{fig:lmc_lr}%
		\end{subfigure}%
		\hfill	
		\begin{subfigure}[b]{0.24\textwidth}
			\includegraphics[width=1.0\textwidth]{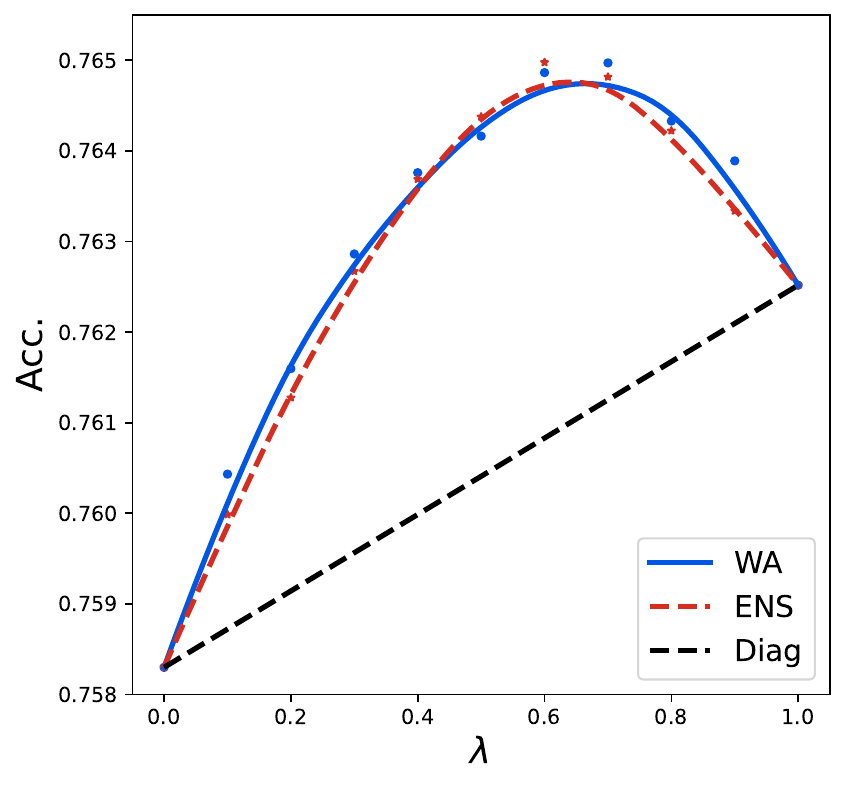}
			\caption{2 RM fine-tunings with different inits: \BAK.}
			\label{fig:lmc_initlr}%
		\end{subfigure}%
	\end{center}%
	\caption{\textbf{Experiments under distribution shifts validating \Cref{obs:lmc,obs:wiens}} on the TL;DR summarization benchmark \cite{volske2017tl}.
	We report the accuracies on $\D_{ood}$ when interpolating between two RM weights $\phi_1$ and $\phi_2$ with the coefficient $\lambda$ sliding between $0$ and $1$.
	\textit{WA} stands for weight averaging $r_{\left(1 - \lambda\right) \cdot \phi_1 + \lambda \cdot \phi_2}$ while \textit{ENS} combines the predictions $(1-\lambda)\times r_{\phi_1} + \lambda \times r_{\phi_2}$; \textit{Diag} is the interpolated accuracy $\left(1 - \lambda\right) \times \acc\left(r_{\phi_1}\right) + \lambda \times \acc\left(r_{\phi_2}\right)$.
	We consider sources of increasing diversity \cite{gontijolopes2022no} between $\phi_1$ and $\phi_2$: in \Cref{fig:lmc_ma8k10k}, they are collected at different number of training steps (8k and 10k) along a single RM fine-tuning; in \Cref{fig:lmc_same}, they are from two independant RM fine-tunings, with the exact same config, but seeing the data in different orders; in \Cref{fig:lmc_lr}, they have different learning rates (1e-4 and 4e-5); in \Cref{fig:lmc_initlr}, they are initalized from different SFT checkpoints collected at different number of SFT steps (8k and 12k), per \BAK introduced in \Cref{fig:baklava}.}%
	\label{fig:lmc_wavsens}%
    \begin{center}
        \begin{subfigure}[b]{0.24\textwidth}
            \includegraphics[width=1.0\textwidth]{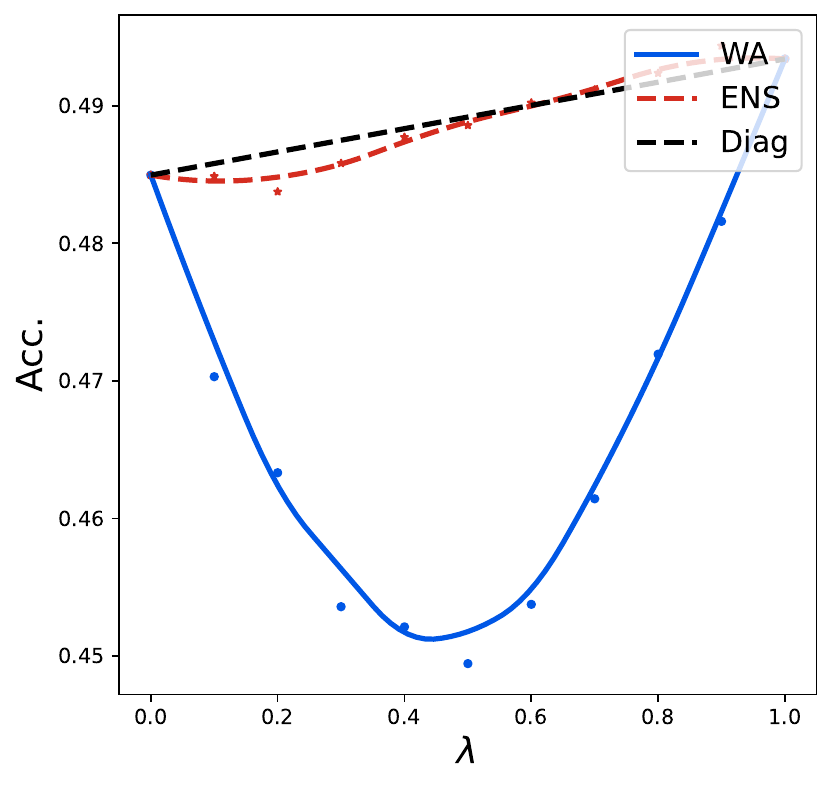}
            \caption{Train (corrupt).}
            \label{fig:corrupt_wavsens_corrupttrain}%
        \end{subfigure}%
        \hfill
        \begin{subfigure}[b]{0.24\textwidth}
            \includegraphics[width=1.0\textwidth]{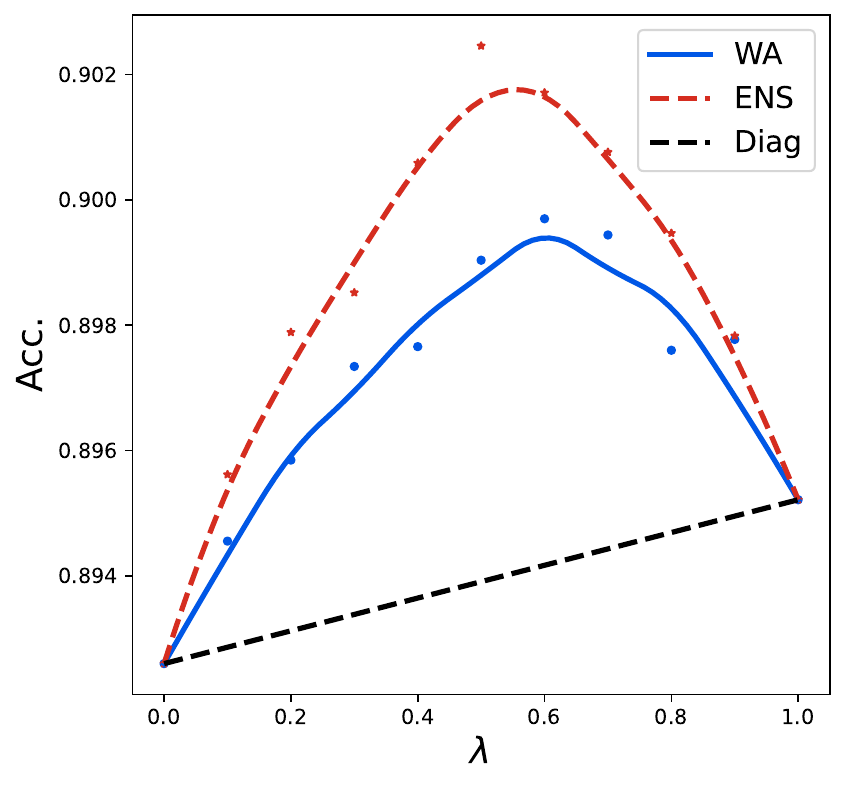}
            \caption{Train (clean).}
            \label{fig:corrupt_wavsens_cleantrain}%
        \end{subfigure}%
        \hfill        
        \begin{subfigure}[b]{0.24\textwidth}
            \includegraphics[width=1.0\textwidth]{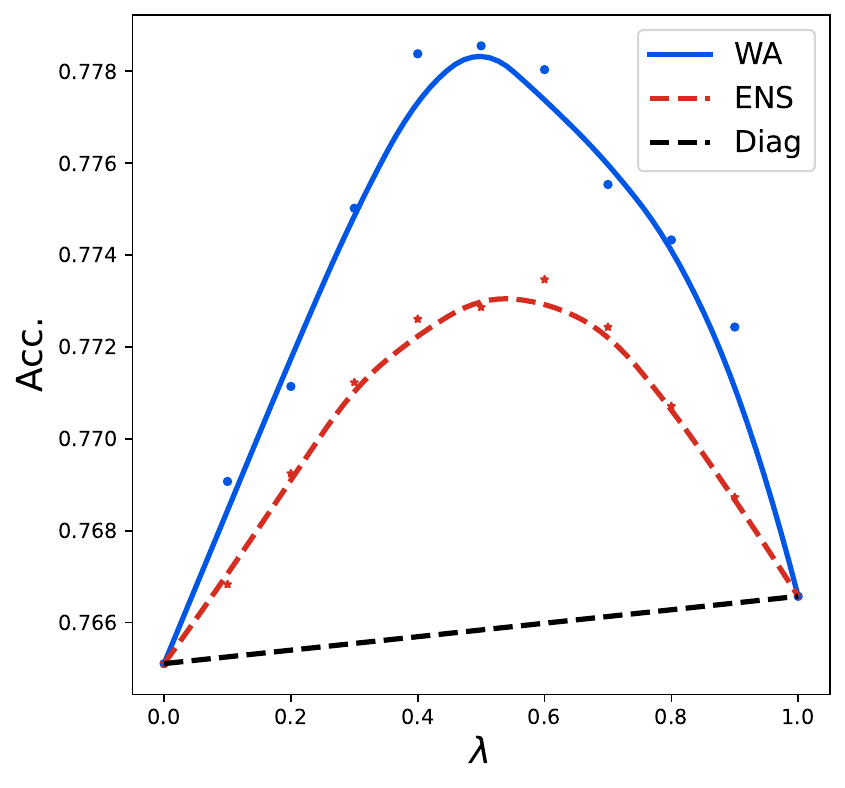}
            \caption{Validation (ID).}
            \label{fig:corrupt_wavsens_valid}%
        \end{subfigure}%
        \hfill        
        \begin{subfigure}[b]{0.24\textwidth}
            \includegraphics[width=1.0\textwidth]{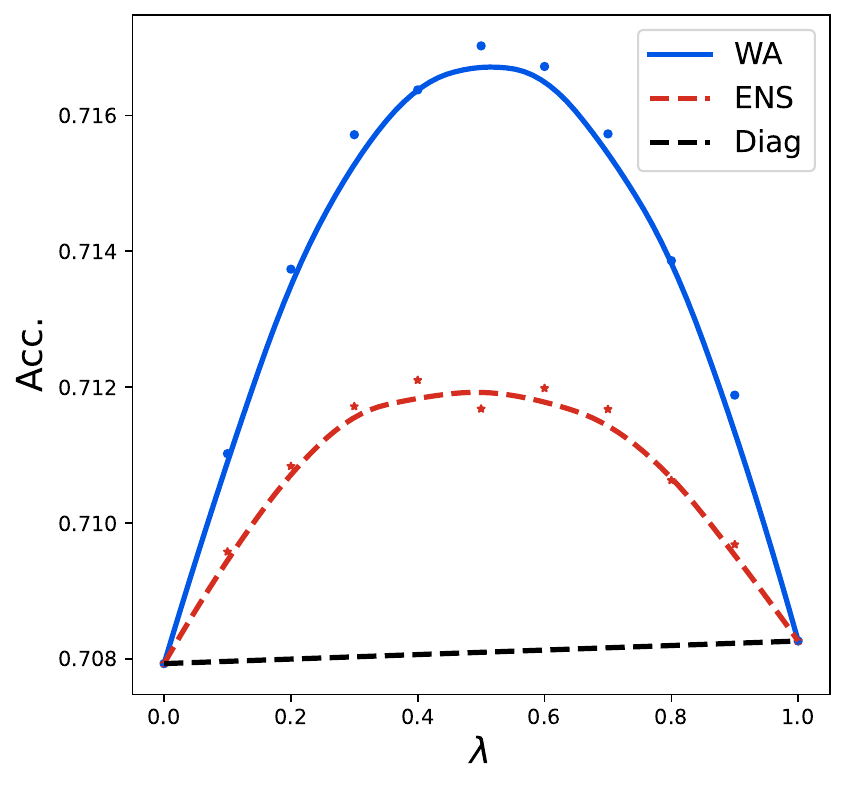}
            \caption{Test (OOD).}
            \label{fig:corrupt_wavsens_ood}%
        \end{subfigure}%
    \end{center}%
    \caption{\textbf{Corruption experiment validating \Cref{obs:wiens2}}. We consider $\phi_1$ and $\phi_2$, two RMs fine-tuned independently with the same config as in \Cref{fig:lmc_same}, but this time with 25\% of the training labels corrupted.
    We then report the performances of their WA and ENS on the different data subsets.
    We observe that WA reduces memorization of the corrupted labels in \Cref{fig:corrupt_wavsens_corrupttrain}, and still performs slightly worse than ENS on the clean training samples in \Cref{fig:corrupt_wavsens_cleantrain}; yet, the performances of WA \wrt ENS improves as we move away from the training distribution, in particular on $\D_{ood}$ in \Cref{fig:corrupt_wavsens_ood} where WA generalizes better.}%
    \label{fig:corrupt_wavsens}%
\end{figure*}%

Theoretically, a simple Taylor expansion can justify this similarity when $\Vert \phi_1 - \phi_2 \Vert \ll 1$.
Empirically, this is validated in \Cref{fig:lmc_wavsens} where the accuracy curves on $\D_{ood}$ for WA and ENS closely match.
This similarity justifies that WA is a variance reduction method; then, because variance is the dominant issue under distribution shifts \cite{rame2022diwa}, this explains the significant gains in \Cref{fig:lmc_wavsens} over the individual RMs $\phi_1$ and $\phi_2$ (validating \Cref{obs:lmc}), in particular when weights are sufficiently diverse.
This suggests improved \reliability in \WARM, with \efficiency benefits over ENS: indeed, WA maintains a single set of weights, removing the memory and inference overheads from ENS.
\subsection{\nth{2} order analysis: weight averaging for more \robust ensembling}
\label{sec:expe:labelnoise}
\textbf{A surprising fact remains unexplained.} WA is slightly superior to ENS under distribution shifts, which one can see on the plots from \Cref{fig:lmc_wavsens}, and more consistently in Figure B.1 from model soups \cite{Wortsman2022ModelSA} or in Figure 1 from DiWA \cite{rame2022diwa}. More generally, WA is the state-of-the-art strategy for OOD generalization, consistently outperforming ENS; yet, this was not explained in previous works, thus urging for new insights about the difference between WA and ENS.%

\textbf{Corruption setup.}
To refine our understanding on the difference between WA and ENS, we propose a new setup where 25\% of the binary labels are swapped in training. We then report the per-subset accuracies on \Cref{fig:corrupt_wavsens}, enriched in \Cref{app:expes:2ndorder} and aggregated in \Cref{fig:wavsens_histogram}.
On the corrupted subset of training data, the accuracy curve for WA is below the expected accuracies, while it is above on all other subsets. More precisely, we make the following \Cref{obs:wiens2}.
\begin{observation}[WA and ENS: \nth{2} order analysis]
	The accuracy gains of WA over ENS grow as data moves away from the training distribution.
	\begin{itemize}
		\item WA $\ll$ ENS on train corrupt: WA is far worse than~ENS on train samples with swapped labels, showing reduced memorization and improved \robustness to label corruption.
		\item WA $\leq$ ENS on train clean: WA is worse than~ENS on train samples with correct labels.
		\item WA $\gtrapprox$ ENS on ID val: WA is better or similar to~ENS on samples without distribution shifts.%
		\item WA $\geq$ ENS on OOD test: WA is far better than~ENS on test samples from new distributions, showing better \reliability under distribution shifts.%
	\end{itemize}%
	\label{obs:wiens2}%
\end{observation}%
\ifthenelse{\boolean{isdouble}}{
\begin{figure}[t!]
	\begin{center}
	\includegraphics[width=0.7\linewidth]{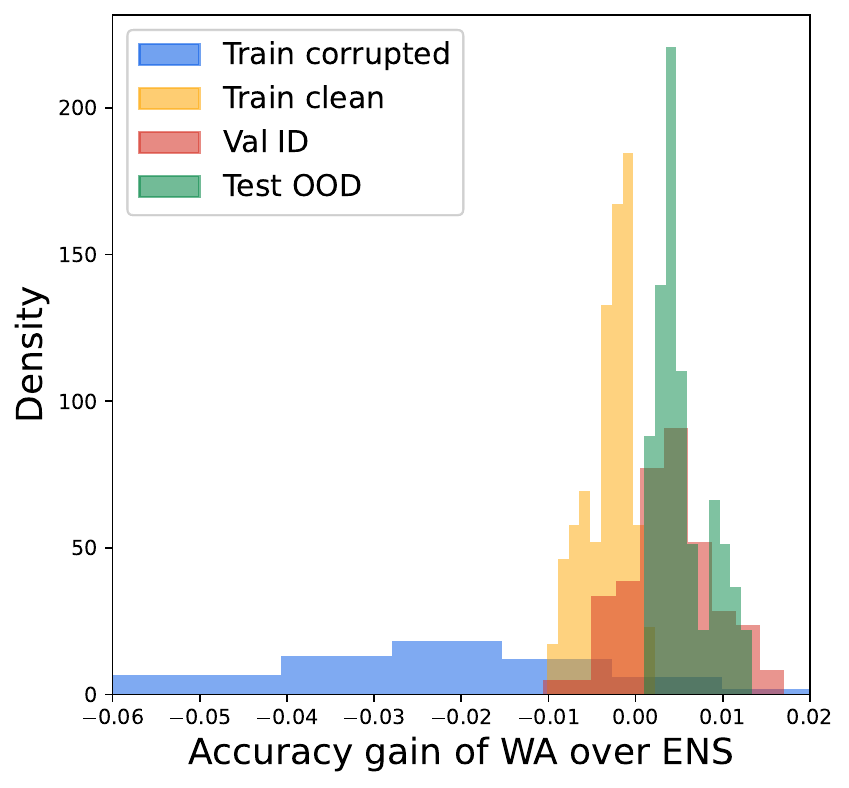}
	\end{center}%
	\caption{Histograms of the differences in accuracy between WA and ENS on different data subsets.}
	\label{fig:wavsens_histogram}
\end{figure}%
}{
\begin{wrapfigure}[10]{hR!}{0.325\textwidth}
	\vspace{-3.5em}
	\centering%
	\includegraphics[width=0.325\textwidth]{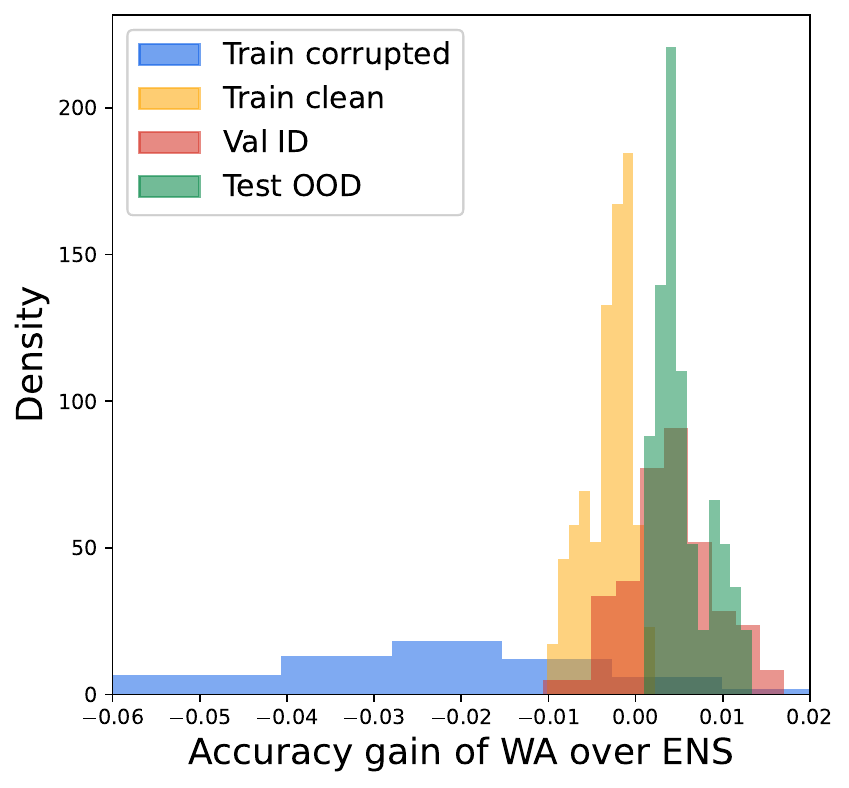}
	\caption{Histograms of the differences in accuracy between WA and ENS on different data subsets.}
	\label{fig:wavsens_histogram}
\end{wrapfigure}%
}
Overall, this suggests that weight averaging memorizes less and generalizes better than ensembling predictions.
\subsection{Weight averaging enforces invariance across runs}
\label{sec:benefits:theory}
We now provide theoretical support to this \Cref{obs:wiens2}.
In brief, our simplifying assumptions suggest that WA acts as a regularization towards the predictive mechanisms that are \textit{invariant} across runs, \ie learned simultaneously in each independent run. Then, in contrast with ENS, WA would improve \robustness to corruption because it would underweight the run-specific features (with low probability of being learned) inducing memorization.

\textbf{Setup.}
We follow Lin \textit{et al.} \cite{lin2023spurious}, and consider a simplified binary classification setup with labels $y\in\{-1,1\}$, related to $F$ features $\{z^j\}_{j=1}^{F}$ such as $z^j\in \sR^d$. From inputs $x$, we train a binary classifier $r(x)=\omega^\intercal f(x)$. Following \cite{lin2023spurious}, we make three key assumptions. First, \textit{features orthogonality}: we assume that $\{z^j\}_{j=1}^{F}$ are orthogonal, \ie $(z^j)^\intercal z^{j'} = 0$ when $j\neq j'$.
Second, \textit{input as bag of features}: we assume that the input $x=\left[x^j\right]_{j=1}^{F} \in \mathbb{R}^{F \times d}$ can be represented as the concatenation of $x^j$ generated by $x^j\sim \gN\left(y \cdot z^j,\sigma \cdot \rmI_d\right)$ with $\sigma \ll 1$. Finally, the \textit{binary featurizer} assumption: we consider that the featurizer $f=\left[f^j\right]_{j=1}^{F} \in \{0,1\}^F$ is a binary selector of the features that make the input. For example, if $y=1$, $F=3$, $x\approx [z^1, z^2, z^3]$, and $f=\left[1, 0, 1\right]$ learns to extract the first and third features, then $f(x)\approx z^1 + z^3$.
We denote $p_j$ the probability that the featurizer $f$ learns to use the $j$-th feature dimension (associated with $z^j$); this means $f^j$ is $1$ with probability $p_j$ and $0$ otherwise.
Moreover, for infinite training samples and under some constraint on $\sigma$, Lemma 5 in \cite{lin2023spurious} proved that, to learn $r=\omega^\intercal f$, the optimal linear fit $\omega$ on the features selected from $f$ would be $\omega=\sum_{j=1}^F f^j \cdot z^j$.

\textbf{Results.}
We consider $M$ RMs $\{r_i=\omega_i^\intercal f_i\}_{i=1}^M$, and compare the limit behaviours of their prediction ensembling $r^{ENS}_M$ and weight averaging $r^{WA}_M$ when $M \to \infty$.
In this limit case, the averaged prediction $r^{ENS}_M=\frac{1}{M}\sum_{i=1}^M \omega_i^\intercal f_i$ for an input $x$ from label $y$ tends towards the expected prediction \linebreak
$\Esp\left[r(x)\right]=\Esp\left[\omega^\intercal f(x) \right]=\Esp_{\{f^j\}_{j=1}^{F}}\big[ \big(\sum_{j=1}^{F} f^j \cdot z^j\big)^\intercal (\sum_{j'=1}^{F} f^{j'} \cdot x^{j'}) \big] \approx y \cdot \sum_{j=1}^{F} p_j \cdot \vert z^j \vert^2$,
using $x^{j'} \approx y \cdot z^{j'}$ thus $(z^j)^\intercal x^{j'} \approx 0$ when $j\neq j'$, and $(f^j)^2=f^j$.%
\begin{equation}
	r^{ENS}_M(x) \xrightarrow[M\to\infty]{} \Esp\left[r(x)\right] \approx y \cdot \sum_{j=1}^{F} \boldsymbol{p_j} \cdot \vert z^j \vert^2.%
\end{equation}%
In contrast, when considering $r^{WA}_M=\left(\frac{1}{M}\sum_{i=1}^M\omega_i\right)^\intercal \left(\frac{1}{M}\sum_{i=1}^Mf_i\right)$ with $M\to\infty$, we have \linebreak$\frac{1}{M}\sum_{i=1}^Mf_i \xrightarrow[M\to\infty]{} \Esp\left[f\right]=\left[p_j\right]_{j=1}^{F}$ and $\frac{1}{M}\sum_{i=1}^M\omega_i \xrightarrow[M\to\infty]{} \Esp\left[\omega\right]=\sum_{j=1}^F p_j \cdot z^j$, and thus:%
\begin{equation}
	r^{WA}_M(x) \xrightarrow[M\to\infty]{} \left(\sum_{j=1}^{F} p_j \cdot z^j\right)^\intercal \left(\sum_{j'=1}^{F} p_{j'} \cdot x^{j'}\right) \approx y \cdot \sum_{j=1}^{F} \boldsymbol{p_j^2} \cdot \vert z^j \vert ^2.%
\end{equation}%

\textbf{Interpretation.}
For ENS, the coefficient for a given feature is $\boldsymbol{p_j}$, the same as the probability of this information being used by any individual network.
In contrast, WA involves the square of the probability $\boldsymbol{p_j^2}$.
Thus WA reduces the reliance on features with low probability, related to minor specific information (such as noise or context) which can be used to fit the corrupted training samples; this would reduce memorization, and thus explains the \robustness of WA under label corruption.
Reciprocally, WA tends to prioritize the most probable features, favoring the mechanisms that are consistently learned, in other words the \textit{mechanisms invariant across runs}.
Overall, WA acts as a regularization, improving \robustness under label corruption by tackling run-specific mechanisms favoring memorization, and improving \reliability under distribution shifts by preserving run-invariant mechanisms favoring generalization.

\begin{remark}[Invariance]
	We argue that weight averaging only keeps the invariant predictive mechanisms across runs. This is in analogy with the invariance literature \cite{muandet2013domain}, popular for domain generalization \cite{arjovsky2019invariant,rame2022fishr} under spurious correlations, where the key idea is that the predictive mechanisms which are \textit{invariant across domains} are the causal ones that are stable under distribution shifts. This theoretically connects two key paradigms for OOD generalization, ensembling and invariance, and shows that weight averaging actually benefits from both.
\end{remark}

\begin{remark}[Extension to a deeper structure with $L$ layers]
We obtain a square in $\boldsymbol{p_j^2}$ due to our simplified two-layer architecture. Yet, in full generality, using a deeper structure with $L$ layers would lead to $\boldsymbol{p_j^L}$. Intuitively, WA applies an AND-mask on the information, that need to be found both in the previous feature space and the next layer weights.%
\end{remark}

\begin{remark}[From reward \robustness to learnability]
When applied to the design of RMs in \WARM, we now argue that WA facilitates \WARM's stability \cite{secret2023rm} by mitigating the reliance on some non-\robust features.
Indeed, WA makes the \WARM reward more \robust to small (potentially adversarial \cite{szegedy2013intriguing}) perturbations \cite{yang2020adversarial}, \ie smoother \cite{rosca2020case} in the input space. This relates to the Lipschitzness property of the reward \cite{hein2017formal,sokolic2017robust,cohen2019certified}, where the difference in predicted rewards is bounded by the distance in input space. Fortunately, such smoothness is useful in RL \cite{hafner2011reinforcement}, in particular for the stability of the policy gradient \cite{pirotta2015policy} because \enquote{sharp changes in reward value are hard to represent and internalize} \cite{blonde2022lipschitzness}.
This is studied in \emph{Lipschitzness is all you need} \cite{blonde2022lipschitzness} where the authors argue that
\enquote{the local Lipschitzness of the reward is a sine qua non condition for good performance}, required \enquote{to even learn anything}.
In summary, \robustness improves stability and hinders the cascade of errors occurring when minor input variations can cause large reward differences.
\end{remark}

\textbf{In conclusion}, we summarize the benefits from \WARM.
First, WARM is \efficient, incurring no memory or computation costs, as it returns a single model.
Second, \WARM reduces variance while leveraging mechanisms invariant across runs, thus improving its \reliability under distribution shifts.
Lastly, \WARM also addresses label corruption, thereby augmenting \robustness to noisy preferences.

\begin{figure*}[b!]
	\begin{center}
		\begin{subfigure}[b]{0.24\textwidth}
			\includegraphics[width=1.0\textwidth]{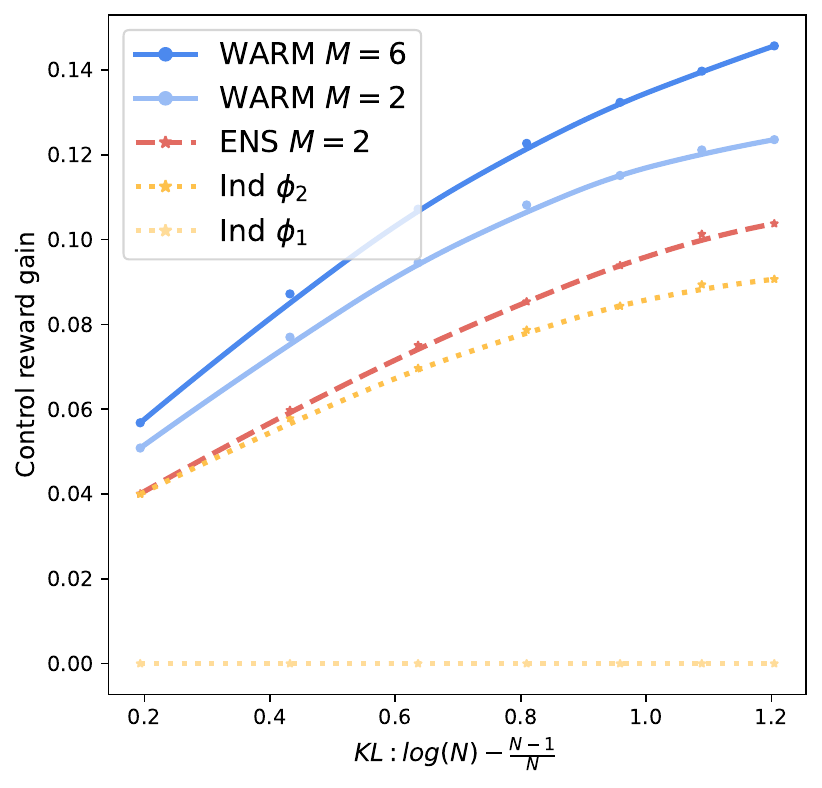}
			\caption{PaLM (clean).}
			\label{fig:bon8_cand_warm_scoregain_selected_kl}%
		\end{subfigure}%
		\hfill
		\begin{subfigure}[b]{0.24\textwidth}
			\includegraphics[width=1.0\textwidth]{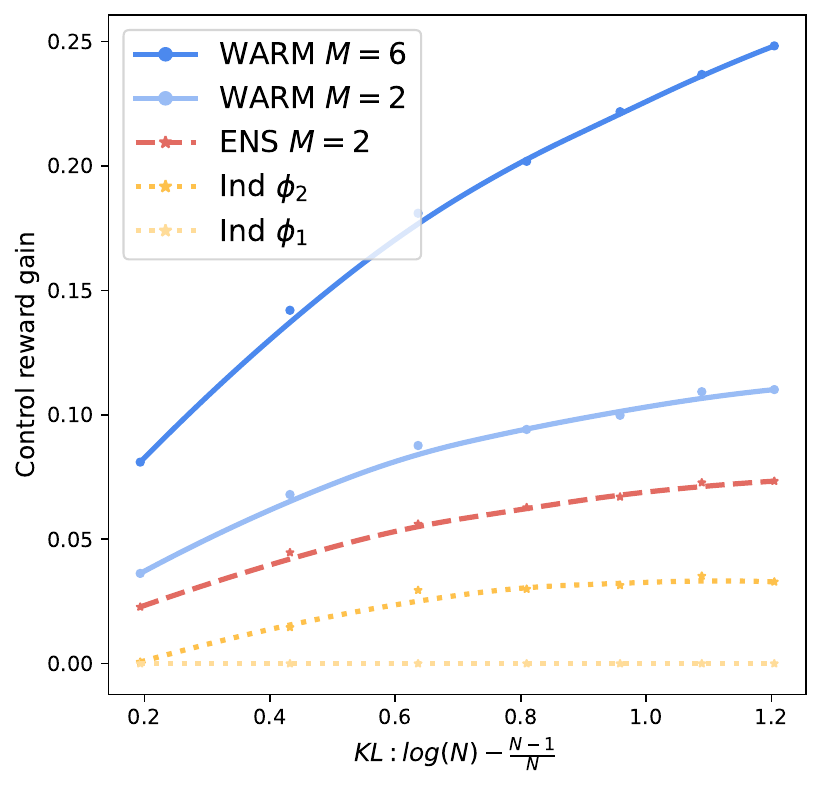}
			\caption{PaLM (corrupt).}
			\label{fig:bon8_cand_warmc_scoregain_kl}%
		\end{subfigure}%
		\hfill
		\begin{subfigure}[b]{0.24\textwidth}
			\includegraphics[width=1.0\textwidth]{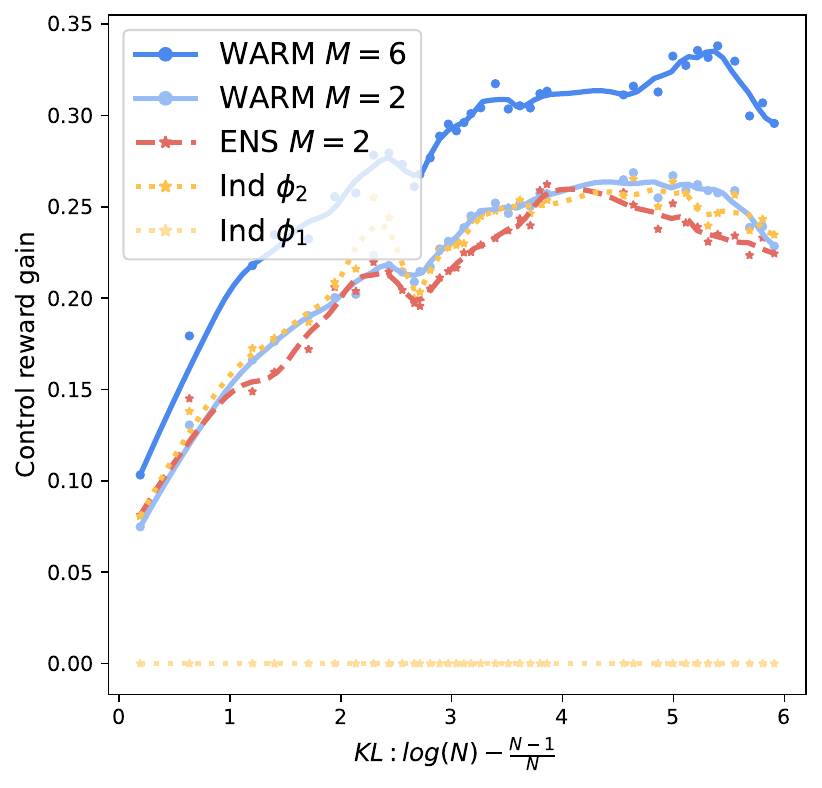}
			\caption{T5 (clean).}
			\label{fig:bon1000_t5x_warm_scoregain_selected_kl}%
		\end{subfigure}%
		\hfill
		\begin{subfigure}[b]{0.24\textwidth}
			\includegraphics[width=1.0\textwidth]{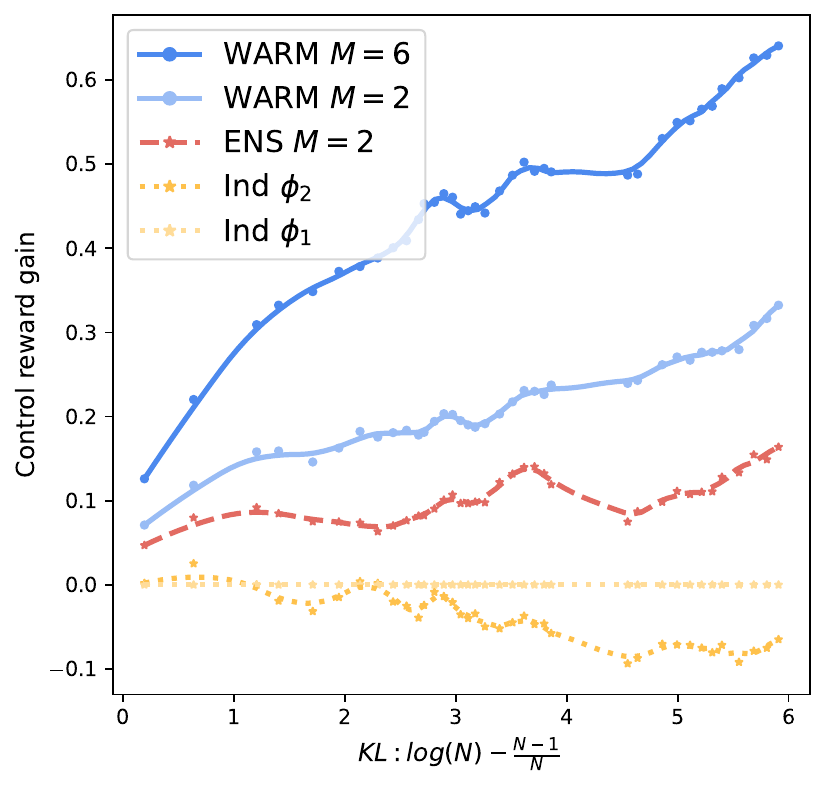}
			\caption{T5 (corrupt).}
			\label{fig:bon1000_t5x_warmc_scoregain_kl}%
		\end{subfigure}%
	\end{center}%
	\caption{\textbf{Control reward for BoN experiments}: clean preference dataset in \Cref{fig:bon8_cand_warm_scoregain_selected_kl,fig:bon1000_t5x_warm_scoregain_selected_kl} and 25\% corruptions in \Cref{fig:bon8_cand_warmc_scoregain_kl,fig:bon1000_t5x_warmc_scoregain_kl}.
	We consider two SFT policies to generate candidate summaries: one based on PaLM architecture \cite{anil2023palm}, the other on T5 architecture \cite{JMLR:v21:20-074}.
	The $x$-axis is the \KL between the BoN policy and the SFT policy; the $y$-axis represents the control reward gains \wrt to an RM $\phi_1$, which was the best individual RM on $\D_{ood}$.
	The blue lines represent \WARM with $M$ weights: \WARM performs higher than the individual RMs (in yellows) or when ensembling their predictions (ENS in red).
	We report the absolute control rewards for those experiments in \Cref{fig:bon_abs}, where the values range roughly between $3$ and $7$.
	}
	\label{fig:bon}%
    \begin{center}
        \begin{subfigure}[b]{0.24\textwidth}
            \includegraphics[width=1\textwidth]{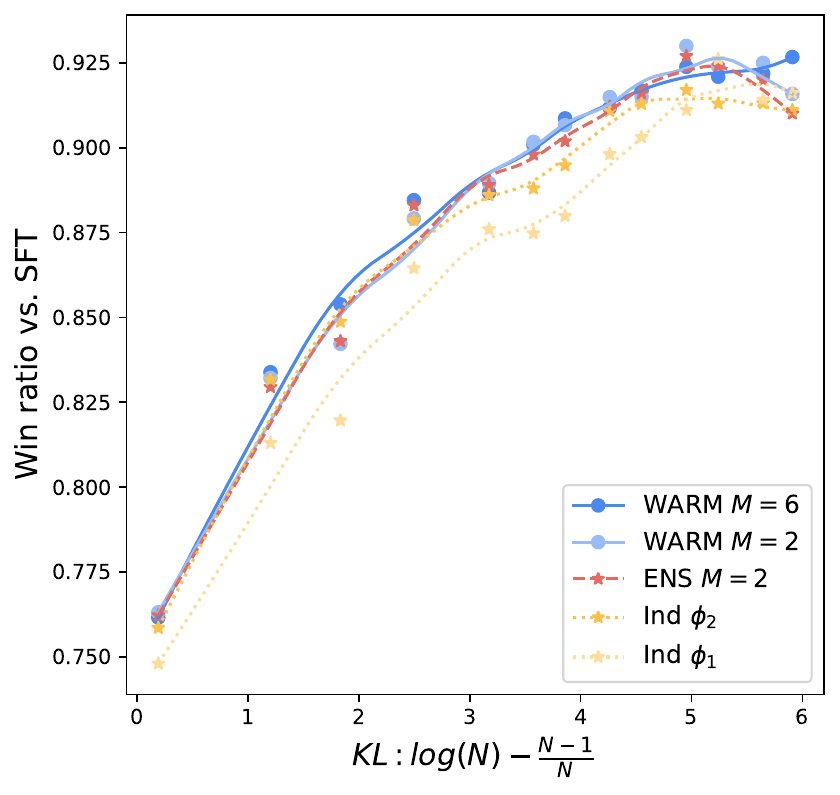}
            \caption{SFT (clean).}
            \label{fig:aipref_bon_t5x_sft}%
        \end{subfigure}%
        \hfill%
        \begin{subfigure}[b]{0.24\textwidth}
            \includegraphics[width=1\textwidth]{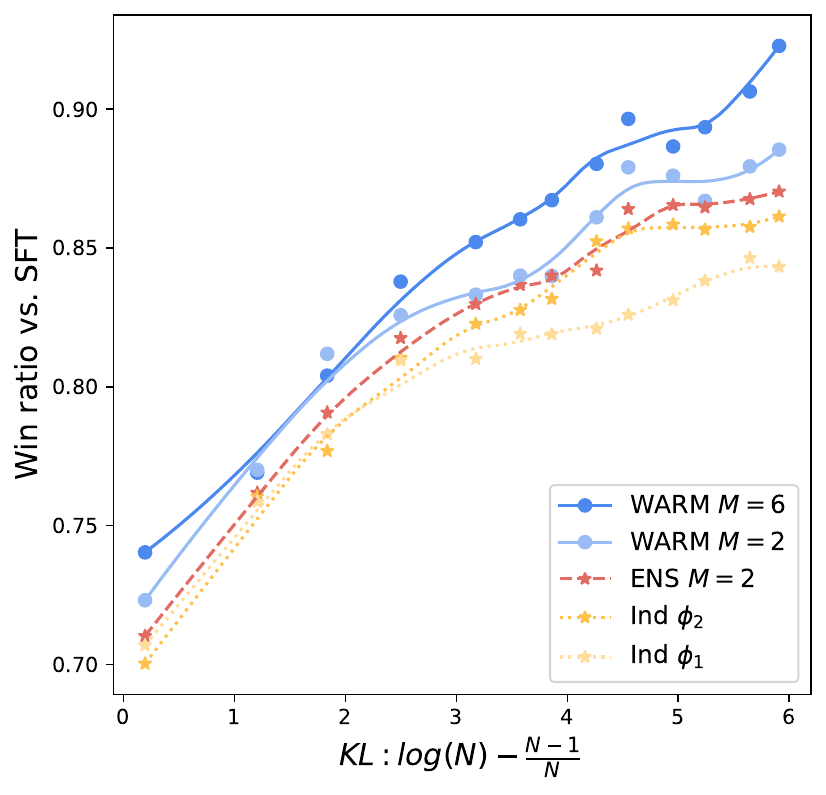}
            \caption{SFT (corrupt).}
            \label{fig:aipref_bonc_t5x_sft}%
        \end{subfigure}%
		\hfill
        \begin{subfigure}[b]{0.24\textwidth}
            \includegraphics[width=1\textwidth]{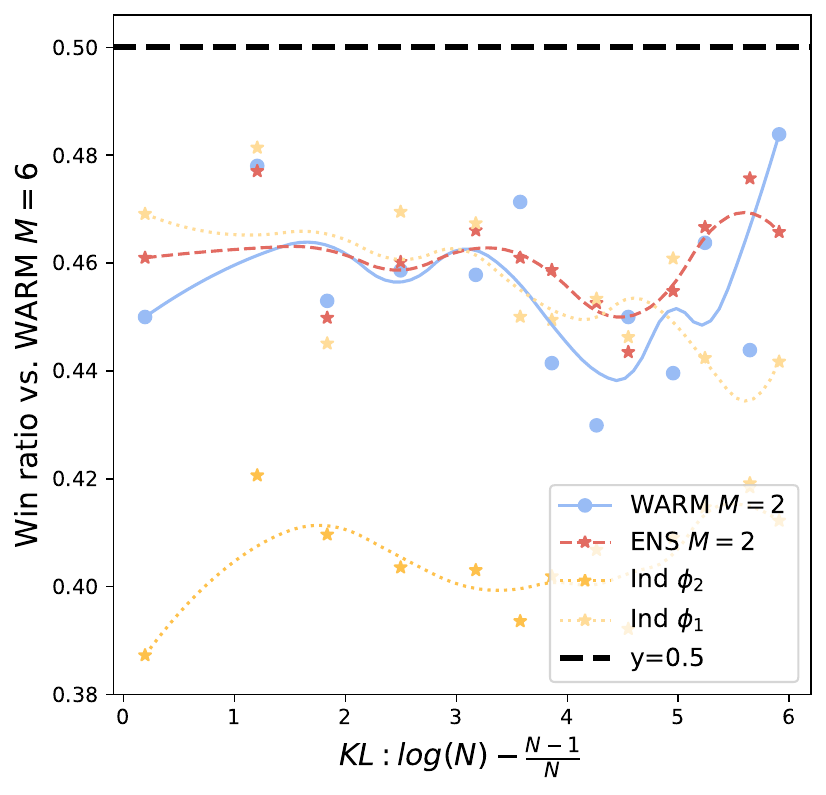}
            \caption{\WARM (clean).}
            \label{fig:aipref_bon_t5x_warm6}%
        \end{subfigure}%
        \hfill%
        \begin{subfigure}[b]{0.24\textwidth}
            \includegraphics[width=1\textwidth]{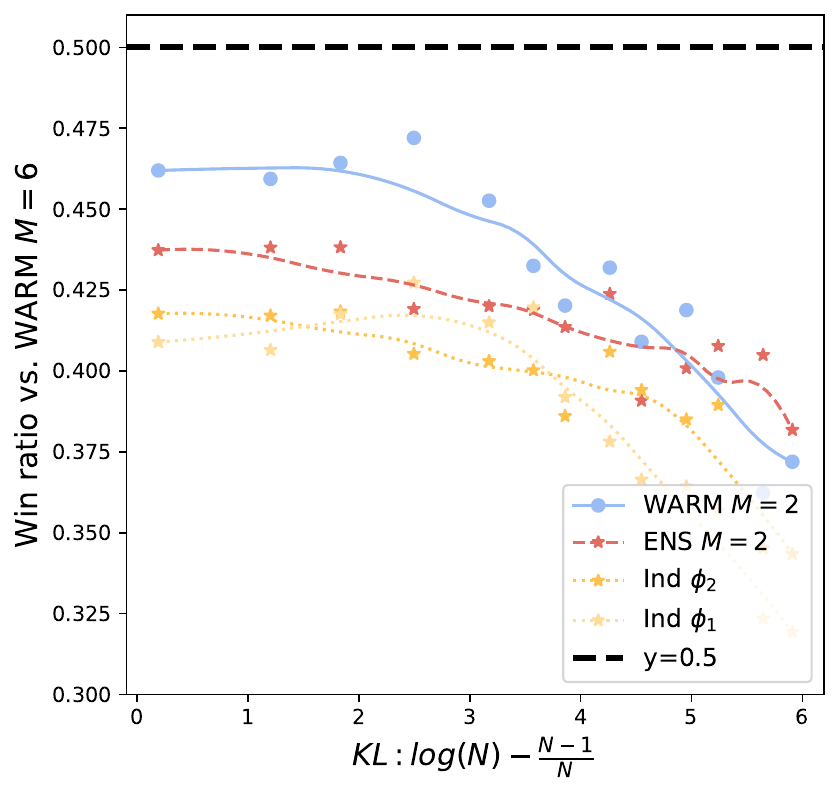}
            \caption{\WARM (corrupt).}
            \label{fig:aipref_bonc_t5x_warm6}%
        \end{subfigure}%
    \end{center}%
    \caption{\textbf{Oracle preference metric for BoN experiments on T5 generations}: clean preference dataset in \Cref{fig:aipref_bon_t5x_sft,fig:aipref_bon_t5x_warm6} and 25\% corruptions in \Cref{fig:aipref_bonc_t5x_sft,fig:aipref_bonc_t5x_warm6}.
	We plot the win rates for different values of $N$ \versus two reference strategies: SFT (\ie random selection or equivalently BoN with $N=1$), or selecting the best summary according to \WARM $M=6$.
	We observe that all strategies beat the SFT reference (they are all above 50\% win rate), but that none beat the \WARM $M=6$ reference.}%
    \label{fig:bon_aipref}%
\end{figure*}%

\section{Experiments}
\label{sec:expe}
To empirically validate \WARM's benefits described in previous section, we train PaLM-XXS RMs on the TL;DR summarization benchmark \cite{volske2017tl} where preference labels are generated by a PaLM-L model prompted with chain-of-thought \cite{wei2022chain}.
This AI labeling approach, increasingly common in recent research \cite{dubois2023alpacafarm,ensemble2023reward,singhal2023long} as an efficient alternative to human assessments, is motivated by studies \cite{bai2022constitutional,lee2023rlaif} indicating that it correlates well with human preferences: critically, it provides an automatic \textit{pairwise oracle preference} metric to evaluate reward hacking (in a similar fashion to the distillation setup from \cite{gao2022scaling}, discussed in \Cref{app:distillation}).
In addition, we leverage a PaLM-XS RM for \textit{pointwise control reward} reaching 80.1\% accuracy on the OOD dataset $\D_{ood}$.
As verified in our experiments, this control RM also detects hacking, as it benefits from a larger architecture and a disjoint pretraining compared to the PaLM-XXS RMs of interest. Below, we explore two key scenarios: in \Cref{sec:expe:bon}, \WARM reranks outputs in best-of-$N$ (BoN); in \Cref{sec:expe:rl}, \WARM guides the RL procedure.
\subsection{Best-of-$N$ experiments}
\label{sec:expe:bon}
\textbf{Setup.}
We start with best-of-$N$ (BoN) sampling experiments in \Cref{fig:bon,fig:bon_aipref}.
Given a dataset of $D$ text prompts, for each prompt we generate $N$ summaries from a SFT policy, and then returns the summary with the highest reward according to different RMs. We actually consider two SFT policies; one based on PaLM architecture~\cite{anil2023palm} ($N=8$, $D=15000$), the other on T5 architecture~\cite{JMLR:v21:20-074} ($N=1000$, $D=1000$).
For the $x$-axis, we plot the \KL between the BoN policy and the SFT policy, which can be approximated by $\log(N)-\frac{N-1}{N}$ \cite{hilton2023kl,beirami2024theoretical}.
BoN is effective \cite{touvron2023llama2}, especially in the low-\KL regime (\ie for small $N$).
We consider two setups, without (clean setup) and with (corrupt setup) 25\% label corruption in the preference datasets for reward modeling, and denote in each setup the weights $\{\phi_i\}_{i=1}^M$ sorted in decreasing accuracy on $\D_{ood}$.

\textbf{Control reward.}
\Cref{fig:bon} shows that, in terms of \textit{pointwise control reward}, \WARM performs consistently better than ENS (only with $M=2$ for computational reasons) and the two best individual RMs $\phi_1$ and $\phi_2$; moreover, the gains get bigger for $M=6$.
As a side note, we also observe that the individual RM $\phi_2$ performs better in BoN in \Cref{fig:bon1000_t5x_warm_scoregain_selected_kl} than $\phi_1$ though $\phi_1$ was better than $\phi_2$ on $\D_{ood}$, highlighting that selecting the appropriate individual RM is not trivial \cite{ensemble2023reward}.

\textbf{Oracle preference.}
In \Cref{fig:bon_aipref}, we leverage the \textit{pairwise oracle preference} \cite{lee2023rlaif} metric to validate better performance with \WARM.
We observe in \Cref{fig:aipref_bon_t5x_sft,fig:aipref_bonc_t5x_sft} that summaries selected with \WARM have a win rate of up to 92.5\% against the random selection of a summary (from SFT).
We also see in \Cref{fig:aipref_bon_t5x_warm6,fig:aipref_bonc_t5x_warm6} that reciprocally, all selection strategies have a win rate lower than 50\% against the summaries selected by \WARM $M=6$.%
\subsection{RL experiments}
\label{sec:expe:rl}
\begin{figure*}[h!]
	\begin{center}
		\begin{subfigure}[b]{0.325\textwidth}%
			\includegraphics[width=\textwidth]{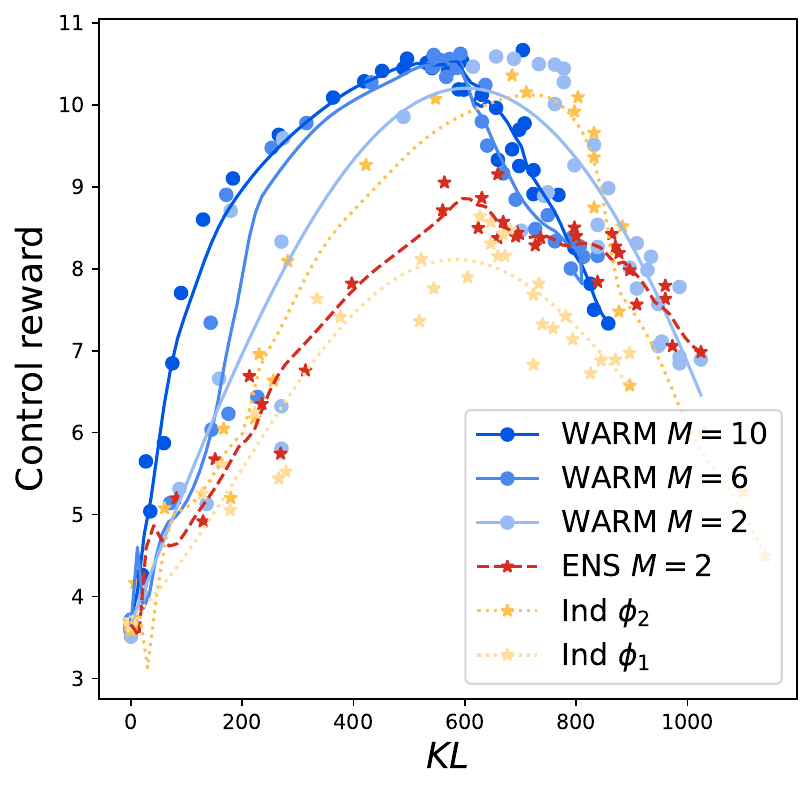}
			\caption{RL (clean).}%
			\label{fig:hackingkl}%
		\end{subfigure}	 
		\hfill
		\begin{subfigure}[b]{0.325\textwidth}
			\includegraphics[width=1.0\textwidth]{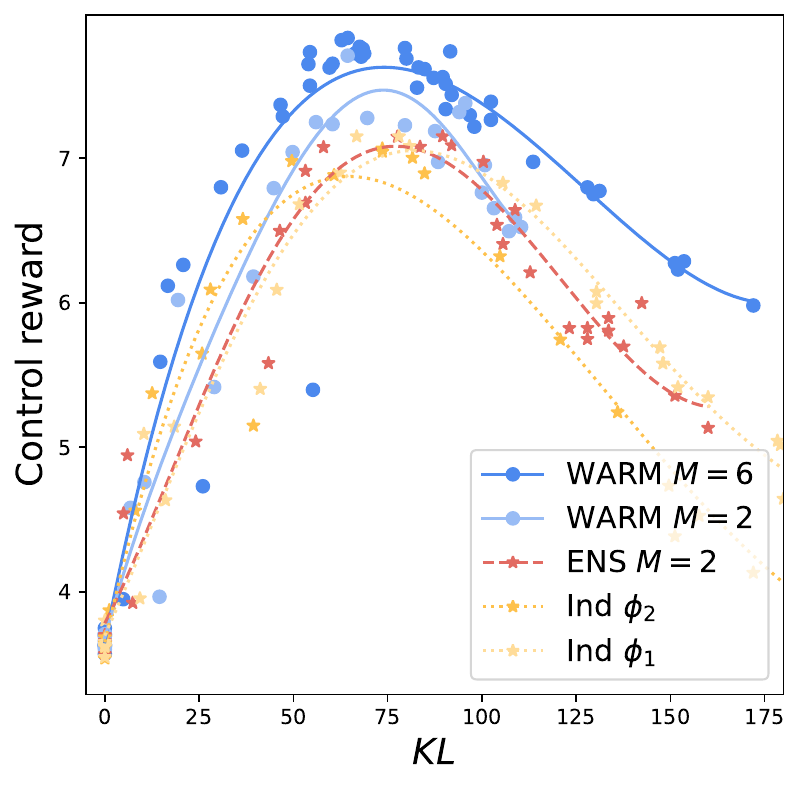}
			\caption{RL (corrupt).}
			\label{fig:hackingkl_corrupt}%
		\end{subfigure}%
		\hfill
		\begin{subfigure}[b]{0.325\textwidth}%
			\includegraphics[width=1.0\textwidth]{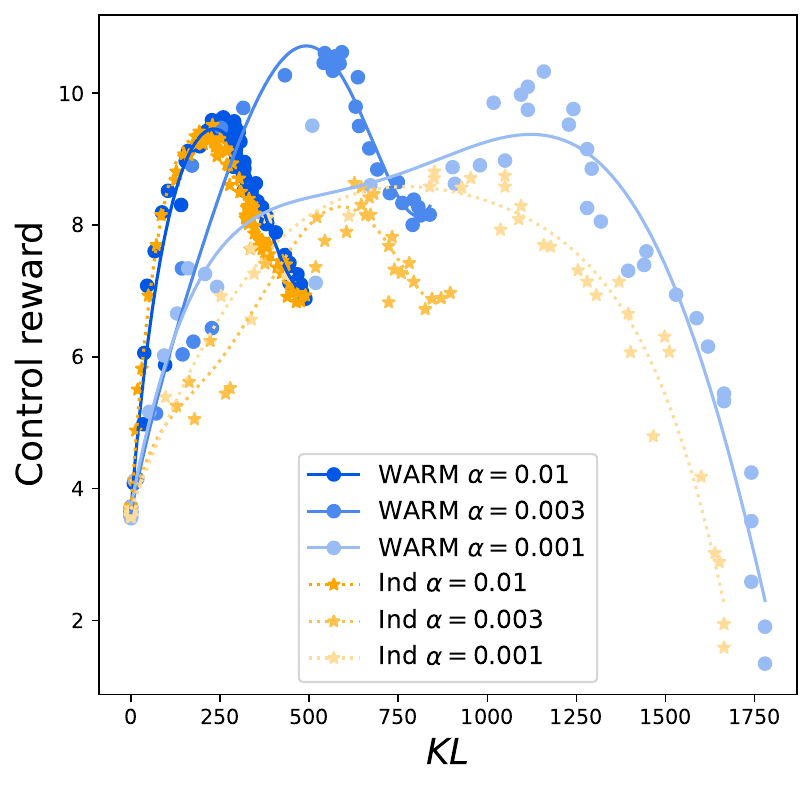}%
			\caption{Ablating $\alpha$ for RL (clean).}%
			\label{fig:controlvskl_alphas}%
		\end{subfigure}		   		
	\end{center}%
	\caption{\textbf{Control reward for RL experiments}: clean preference dataset in \Cref{fig:hackingkl,fig:controlvskl_alphas} and 25\% corruptions in \Cref{fig:hackingkl_corrupt}.
	The blue lines show the RL fine-tuning of policies when averaging $M$ weights as the RM; the darker, the higher the $M$.
	It performs higher than when RL fine-tuning with the individual RMs (in yellows) or when ensembling their predictions (in red).
	\Cref{fig:controlvskl_alphas} shows results of policies RL fine-tuned with \WARM $M=6$ or $\phi_1$, for different values of $\alpha$ controlling the \KL regularization strength.
	}%
	\label{fig:controlvskl_hacking}%
    \begin{center}
        \begin{subfigure}[b]{0.325\textwidth}
            \includegraphics[width=1\textwidth]{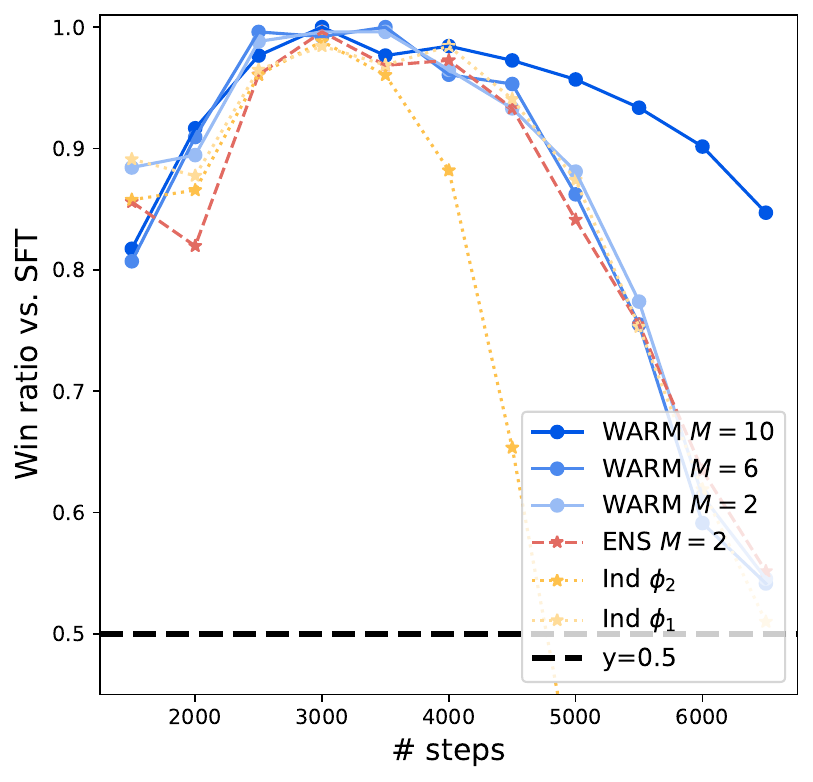}
            \caption{SFT.}
            \label{fig:aipref_sftvsall_alpha003}%
        \end{subfigure}%
        \hfill%
        \begin{subfigure}[b]{0.325\textwidth}
            \includegraphics[width=1\textwidth]{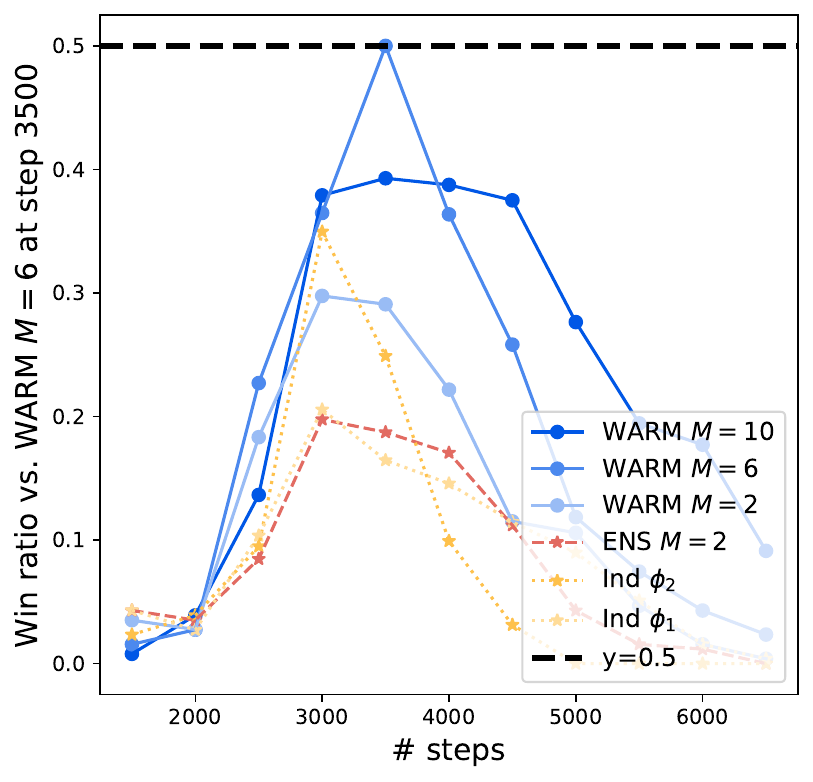}
            \caption{WARM $M=6$.}
            \label{fig:aipref_warm63500vsall_alpha003}%
        \end{subfigure}%
        \hfill%
        \begin{subfigure}[b]{0.325\textwidth}
            \includegraphics[width=1\textwidth]{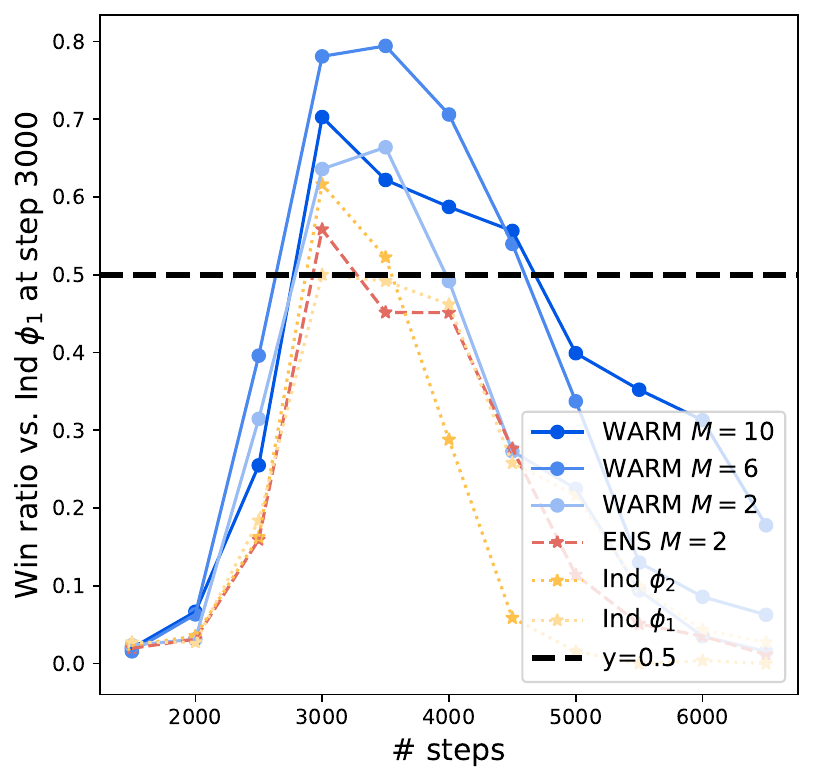}
            \caption{$\phi_1$ (best individual RM).}
            \label{fig:aipref_lr4e53000vsall_alpha003}%
        \end{subfigure}%
        \hfill%
    \end{center}%
    \caption{\textbf{Oracle preference metric for RL experiments}: clean preference dataset. We plot the win rates along RL fine-tuning against three reference policies: the SFT policy, the policy RL fine-tuned with \WARM $M=6$ after 3500 steps, and the policy RL fine-tuned with $\phi_1$ after 3000 steps. \Cref{fig:rl_aipref_fixedepochs} reports results when comparing policies at fixed number of training steps.}%
    \label{fig:rl_aipref}%
\end{figure*}%

\textbf{Setup.}
For RL fine-tuning of policies, we follow \cite{lee2023rlaif} and use their modified version of REINFORCE~\cite{williams1992simple} with a baseline value score for variance reduction, a simpler algorithm than PPO~\cite{schulman2017proximal} yet still effective for LLMs.
Both policy and value LLMs are PaLM-XS, initialized from the same SFT model. We then generate samples with the policy, compute the reward with the RMs and update the weights to optimize this reward. More details are available in \Cref{app:details:rl}.
To reduce forgetting and encourage the policy to remain close to its SFT initialization, we incorporate a \KL regularization \cite{jaques2017sequence,geist2019theory} controlled by a coefficient $\alpha$,  ablated in \Cref{fig:controlvskl_alphas}, yet otherwise set to $0.003$ in the clean setup and $0.01$ in the corrupt setup.
This \KL serves as the $x$-axis in our plots to estimate model drift, as done in the literature; same curves with the number of training steps as the $x$-axis in \Cref{fig:main:hacking,fig:controlvsstep_corruptalp}.

\textbf{Control reward.}
In \Cref{fig:controlvskl_hacking}, we observe reward hacking; as the policy moves away from its SFT initialization, the control reward collapses.
Critically, \WARM improves performances: in particular, increasing $M$ pushes the Pareto front of solutions to the top left in \Cref{fig:hackingkl,fig:hackingkl_corrupt}.
In comparison, policies trained with ENS (with $M=2$ for computational reasons) are still susceptible to early reward hacking, while reaching absolute control rewards significantly worse than with \WARM (even with $M=2$).
In \Cref{fig:controlvskl_alphas}, we confirm that the $\alpha$ hyperparameter plays a crucial role; low values of $\alpha$ such as $0.001$ correspond to high \KL, while high values of $\alpha$ such as $0.01$ entail low \KL but a risk of underfitting.
From a practical perspective, this highlights that the optimal value of $\alpha$ for \WARM is lower than for a single RM; this is because \WARM can mitigate reward hacking, and thus the optimal policies are obtained for larger values of \KL.
\newpage
\textbf{Oracle preference.}
In \Cref{fig:rl_aipref}, we compare the different policies according to our pairwise oracle preference AI labeler \cite{lee2023rlaif}.
In \Cref{fig:aipref_sftvsall_alpha003}, the reference policy is the SFT initialization; all the RL fine-tuned policies outperform this baseline, with \WARM $M=6$ reaching a win rate of $99.8\%$ after 3500 steps (the highest win rate among all policies).
We use this policy as the reference in \Cref{fig:aipref_warm63500vsall_alpha003}; no other policy could beat it.
Interestingly, we observe that using $M=10$ rewards can delay reward hacking but does not improve the peak performance; we speculate this is related to our weight selection procedure, as the weights $\{\phi_i\}_{i=7}^{10}$ have lower individual accuracy on $\D_{ood}$ than $\{\phi_i\}_{i=1}^6$ (more details in \Cref{fig:weight_selection}).
Finally, in \Cref{fig:aipref_lr4e53000vsall_alpha003}, the reference policy is obtained after 3000 steps of RL fine-tuning with $\phi_1$ (the best individual RM on $\D_{ood}$). There is a large region of steps in which policies trained \WARM (even with $M=2$) beat this approach; the previous reference from \Cref{fig:aipref_warm63500vsall_alpha003} actually has a 79.4\% win rate against it.

\newpage
\section{Discussion}%
\label{sec:limitations}%

\textbf{Benefits.}
\WARM represents a flexible and pragmatic method to improve the alignment of AI with human values and societal norms. This paper has detailed several of its benefits, and below, we delve into additional, more exploratory advantages.
\WARM follows the \textit{updatable machine learning paradigm} \cite{updatablemachinelearning}, eliminating the need for inter-server communication, thus enabling \textit{embarrassingly simple parallelization} \cite{li2022branch} of RMs.
This facilitates its use in \textit{federated learning} scenario \cite{mcmahan2017communication} where the data should remain private; moreover, WA would add a layer of privacy and bias mitigation by reducing the memorization of private preference \cite{zaman2023fuse}.
Then, a straightforward extension of \WARM would combine RMs trained on different datasets, for example, coming from different (clusters of) labelers.
This diversity could help \WARM performances, but also from a multi objective perspective \cite{wu2023finegrained}; by non-uniform interpolation of RMs, we could learn a set of \textit{personalized policies} \cite{rame2023rewarded}.
Furthermore, as WA has been shown to limit catastrophic forgetting \cite{stojanovski2022momentum,eeckt2022weight}, \WARM could seamlessly support iterative and evolving preferences.
Finally, a promising research direction is extending \WARM to direct preference optimization (DPO) strategies \cite{rafailov2023direct}, where averaging the RMs casts back to averaging the DPO policies \cite{neuralbeagle24}.

\textbf{Limitations.}
\WARM, while innovative, does face some limitations, notably two when compared to prediction ensembling methods; first, prediction ensembling can benefit from the diversity brought by combining RMs from various architectures and pre-trainings; second, prediction ensembling can incorporate prediction disagreement into the reward to provide uncertainty estimation and limit model drift.
However, it's been noted in  \cite{ensemble2023reward} that simple averaging of logits often performs comparably to more complex prediction aggregation functions that include uncertainty elements.
Another limitation is that, while \WARM effectively reduces certain types of memorization, it does not completely eradicate all forms of spurious correlations or biases inherent in the preference data. For instance, if each individual RM predominantly relies on summary length as a criterion, \WARM is likely to replicate this tendency.
Therefore, alternative methods (from the OOD generalization literature?) might be required, for example those based on invariance regularization \cite{arjovsky2019invariant,rame2022fishr} or last layer retraining \cite{kirichenko2022last}.
Finally, \WARM only enhances reward modeling without tackling the other challenges in RLHF \cite{casper2023open}; thus, to mitigate the safety risks~\cite{amodei2016concrete,hendrycks2022x,hendrycks2023natural} from misalignment~\cite{taylor2016alignment,ngo2022alignment}, \WARM must be considered within the larger context of responsible AI.

\section{Conclusion}%
In conclusion, we introduce Weight Averaged Reward Models (\WARM) to address two critical challenges in reward modeling: \reliability under distribution shifts and \robustness under label corruption.
By averaging the weights of multiple RMs obtained from diverse fine-tunings, \WARM appears as an \efficient solution to mitigate reward hacking in reinforcement learning from human feedback.
Our empirical results demonstrate its effectiveness when applied to summarization. We anticipate that \WARM will contribute to more aligned, transparent, and effective AI systems, encouraging further exploration in reward modeling.

\clearpage
\newpage
\bibliographystyle{unsrt}
\bibliography{main}

\clearpage
\newpage
\appendix
\hrule
\begin{center}
    \Large \WARM: On the Benefits of Weight Averaged Reward Models
\end{center}

\begin{center}
    \large Supplementary material
\end{center}
\hrule
\vskip 0.5cm
This supplementary material is organized as follows:
\begin{itemize}
    \item \Cref{sec:related} enriches our related work.
    \item \Cref{app:details} clarifies some experimental details.
    \item \Cref{app:expes} enriches our experiments.
\end{itemize}

\section{Related work}
\label{sec:related}

This paper leverages the insights from the OOD generalization literature, in particular from linear mode connectivity (see \Cref{sec:related:wa}), and applies them to the design of \efficient, \reliable and \robust reward models (see \Cref{sec:related:rm}).
\subsection{Out-of-distribution generalization, linear mode connectivity and memorization}
\label{sec:related:wa}

\textbf{LMC in fine-tuning.} Fine-tuning foundation models \cite{bommasani2021opportunities} into specialized models that generalize well to new distributions is critical for many real-world applications \cite{hendrycks2018benchmarking,Zech2018,degrave2021ai}.
Recently, different variants of weight averaging (WA) were able to improve performance, such as moving average~\cite{izmailov2018,cha2021wad,arpit2021ensemble}, WiSE fine-tuning~\cite{Wortsman2022robust}, model soups~\cite{Wortsman2022ModelSA}, DiWA~\cite{rame2022diwa} and model ratatouille \cite{rame2022recycling}.
These works rely on the LMC~\cite{Frankle2020,Neyshabur2020} across fine-tuned weights, which was extended to fine-tunings on different tasks \cite{ilharco2022patching,choshen2022cold,rame2022recycling}, modalities \cite{mustafaunival2023} or with different losses \cite{rame2022diwa,croce2023seasoning}, although \cite{juneja2022linear} highlighted some limitations.
WA was also used recently in RL setups \cite{nikishin2018improving,gaya2021learning,lawson2023merging,rame2023rewarded,noukhovitch2023language}, in particular in RLHF in \cite{rame2023rewarded,noukhovitch2023language} but only to combine policies and not rewards.

\textbf{Insights into WA.}
Specifically, WA comes with several benefits.
First, WA flattens the loss landscape \cite{cha2021wad}.
Second, WA approximates prediction ensembling, thus reduces variance of the estimator \cite{Wortsman2022ModelSA,rame2022diwa} and tackles model misspecification \cite{d2020underspecification}.
Third, WA combines models' abilities \cite{2022arXiv221204089I,daheim2023elastic}, which can be useful for multi-task \cite{ilharco2022patching}, multi-objective \cite{rame2023rewarded} or in continual learning \cite{stojanovski2022momentum} setups.
Lastly, it has recently been shown that WA can provide some benefits under spurious correlations \cite{lin2023spurious,zaman2023fuse}, with a phenomenon called \emph{FalseFalseTrue} in \cite{lin2023spurious}. These works \cite{lin2023spurious,zaman2023fuse} share similarities with our memorization experiments from \Cref{sec:expe:labelnoise}, but we are the first to analyze WA regularization properties under label corruption, and their consequences on generalization.
In contrast, in \cite{zaman2023fuse} the networks are trained on different datasets while the theory in \cite{lin2023spurious} is actually mostly developed for prediction ensembling.

\textbf{Memorization.}
Traditional approaches \cite{song2022survey} tackling memorization of corrupted labels \cite{zhang45820} usually require explicit regularization \cite{tanno2019learning}, specific data augmentation \cite{jain2023neftune}, loss adjustment \cite{ghosh2017robust} or sample selection \cite{xia2022sample}.
Some other strategies are based on ensembling: they filter out potentially corrupted samples with self-labeling filtering \cite{jiang2018mentornet,han2018co} or bagging diversity procedures \cite{sabzevari2019ensemble}. As far as we know, with WA we propose the first strategy combining multiple models trained on the same dataset that manages to tackle corruption.

\subsection{Reward modeling}
\label{sec:related:rm}
One of the central challenge in aligning LLMs is the absence of explicit rewards from the environment, \aka the outer alignment challenge \cite{ngo2022alignment}.
While Inverse Reinforcement Learning \cite{ng2000algorithms} attempts to derive a reward model (RM) from expert demonstrations, most recent efforts \cite{christiano2017deep,ziegler2019fine,stiennon2020learning,wu2021recursively,ouyang2022training} primarily focus on learning from human preferences.
Despite its importance to enhance LLM performances post-RL and for safe deployment in real-world applications, how to best design RMs has arguably receive less attention than it warrants.
Some research \cite{knox2023learning} seeks to refine the loss function from \Cref{eq:rm}.
Other approaches are more data oriented: for example, LLaMA-2 \cite{touvron2023llama2} involves continual learning of the RM to adjust to new generation distributions; \cite{reddy2020learning,barnett2023active} follow an active learning paradigm \cite{gooding2023impact}.
Augmenting rewards with tools \cite{li2023tool} or additional information \cite{sun2023aligning} represents an even more recent and very promising trend.
Limited efforts have been made at the intersection of label corruption and reward modeling; \cite{anonymous2023rime} tried to filter the preference dataset for small academic locomotion tasks, while the concurrent \cite{secret2023rm} suggests applying label smoothing and flipping.
Actually, reward ensembling is the most discussed method to mitigate reward hacking \cite{coste2023reward,ensemble2023reward}; we show that \WARM can beat ENS while removing its overheads.
Finally, following DPO \cite{rafailov2023direct}, a recent trend merges reward modeling with policy learning; though, the policies still tend to hack the preference data \cite{azar2023general}, and thus require only a few training steps and very small learning rates.
The WA of DPO policies, theoretically equivalent to the WA of RMs, is a promising research direction with already significant empirical results on public benchmarks, as demonstrated in \cite{neuralbeagle24}.

\section{Implementation details}
\label{app:details}

\subsection{Dataset details}

For summarization, we use the Reddit TL;DR dataset~\cite{stiennon2020learning}, containing posts from Reddit that have been filtered to ensure high quality.
The training summaries from \cite{stiennon2020learning} are generated by OpenAI GPT-3 \cite{NEURIPS2020_1457c0d6} variants.
The dataset contains 123k posts, and $\sim$5\% is held out as the ID validation set.
To generate the candidate responses in the OOD dataset $\D_{ood}$ with 92k pairwise comparisons, we considered multiple PaLM-XS policies with high temperature, some of which are pre-trained only, others SFT-ed and others RLHF-ed; the goal was to get a diverse set of summaries.
\subsection{AI labeling details}
\label{app:details:aipref}
While the ideal approach for evaluating our models would involve human preferences,
we resort to the cheaper AI labeling procedure from RLAIF \cite{lee2023rlaif}.
We query an instruct fine-tuned PaLM-L~\cite{anil2023palm} LLM\footnote{Available through Google Cloud's Vertex AI \url{https://cloud.google.com/vertex-ai/docs/generative-ai/learn/models}.}, prompted to generate preference mimicking human preferences.
Specifically, we follow the \enquote{Detailed + CoT 0-shot} prompting strategy from RLAIF \cite{lee2023rlaif}, the best one according to their results, involving zero-shot prompting with chain-of-thought \cite{wei2022chain}, a maximum decoding length of 512 tokens and temperature $T = 0.0$ (\ie greedy decoding).
To avoid position bias, we run the AI labeler in the two possible orderings.
This strategy was shown to perform similarly to human labellers, with similar inter-agreement.
For the corruption experiments, we swap the labels for 25\% of the training samples.

\subsection{Reward modeling details}
\label{app:details:rm}
The RMs are PaLM-XXS models \cite{anil2023palm}.
They are first pre-trained, and then supervised fine-tuned on the Reddit TL;DR dataset for 12k steps with a batch size of 128 and the Adafactor~\cite{shazeer2018adafactor} optimizer with a learning rate of $10^{-5}$.
Following the \BAK recipe, we actually launch the reward modeling from different checkpoints along this SFT fine-tuning, at steps $\{$8k, 10k, 12k$\}$; taking a too-early checkpoint would drastically reduce RM accuracy, as observed in \cite{razin2023vanishing}.
To convert this LLM into a classifier, we plug a linear probed \cite{kumar2022finetuning} classification layer (the same for all RMs); said differently, even though the featurizers are actually from different SFT checkpoints, they share the same linear probed classification linear layer. As explained in \cite{kumar2022finetuning}, it prevents features from moving too much away from their initializations, which facilitates the LMC required for WA.

We train all RMs for 10k steps, a batch size of 128, the Adafactor~\cite{shazeer2018adafactor} optimizer, a learning rate sampled in $\{$1e-5,4e-5,1e-4$\}$, and a dropout probability in $\{$0.05, 0.1$\}$.
This follows the practical recommandations from \cite{rame2022diwa} to leverage hyperparameters in a mild range to preserve the LMC.
Training for a longer number of steps could help, as it did not alter the LMC in previous works \cite{rame2022recycling}.

\begin{figure}[h!]
	\begin{center}
		\includegraphics[width=0.325\textwidth]{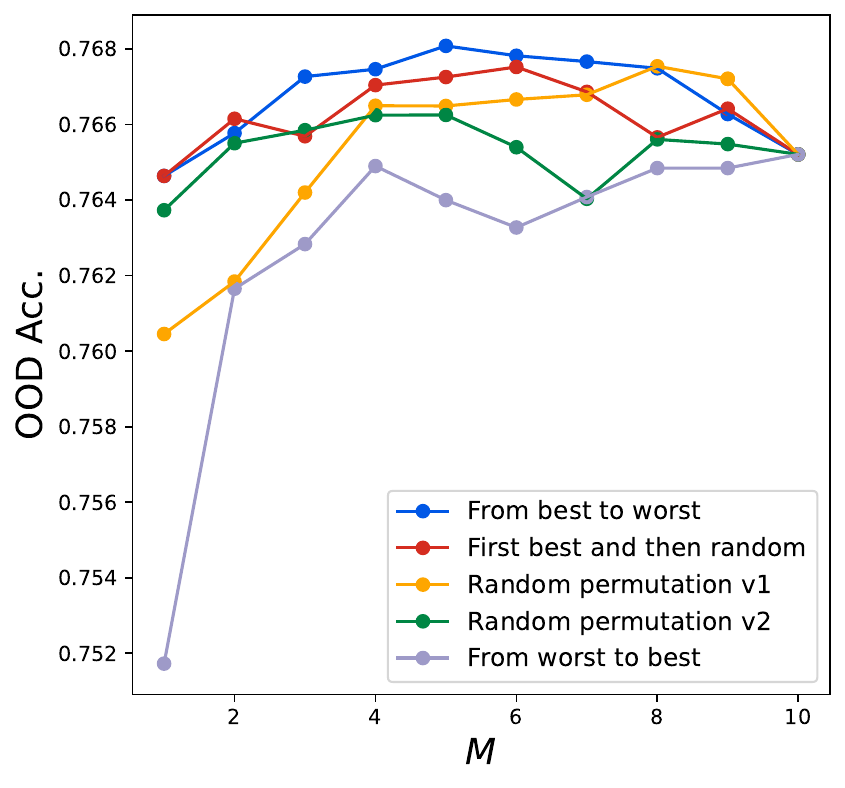}
        \caption{\textbf{Analysis of the weight selection procedure}.
		We plot the accuracy resulting from averaging $M$ weights (out of 10), where these weights are chosen based on various selection procedures. This effectively validates that choosing models from best to worst serves as a reliable heuristic.}%
        \label{fig:weight_selection}%
	\end{center}%
\end{figure}

In practice, for the main experiments with clean labels, we launch 10 reward modelings; when ranked in decreasing accuracy on $\D_{ood}$, we denote them $\{\phi_i\}_{i=1}^{10}$.
Therefore, the RMs named $\phi_1$ and $\phi_2$ in the different plots are the two best according to their individual performances under distribution shifts.
Then, \WARM $M=2$ is actually the RM defined per $\frac{\phi_1+\phi_2}{2}$, while ENS $M=2$ averages their predictions. More generally, \WARM with $M$ weights is the WA of the $M$ best weights $\{\phi_i\}_{i=1}^{M}$.
The main motivation of this weight selection procedure is to remove potentially bad RMs, as validated in \Cref{fig:weight_selection}, in which we consider different permutations across those 10 RMs. As a side note, we speculate that a greedy procedure as in \cite{Wortsman2022ModelSA} could further improve performances.

\FloatBarrier
\subsection{Reinforcement learning details}
\label{app:details:rl}
Both policy and value models are PaLM-XS \cite{anil2023palm}, initialized from the same SFT model. We then generate samples from the policy with temperature $T=0.9$, batch size of 128, the Adafactor~\cite{shazeer2018adafactor} optimizer, a learning rate of $10^{-5}$ and a policy warmup of 2k steps. We set $\alpha=0.003$ for the \KL regularization in the main experiment without label corruption, and $\alpha=0.01$ with label corruption. Following \cite{lee2023rlaif}, we used a modified version of REINFORCE~\cite{williams1992simple} with a baseline value function for variance reduction.

\FloatBarrier
\newpage
\section{Additional experiments}
\label{app:expes}
\subsection{\nth{2} order analysis: weight averaging for more \robust ensembling}
\label{app:expes:2ndorder}
\FloatBarrier
\begin{figure}[h!]
	\begin{center}
		\includegraphics[width=0.55\textwidth]{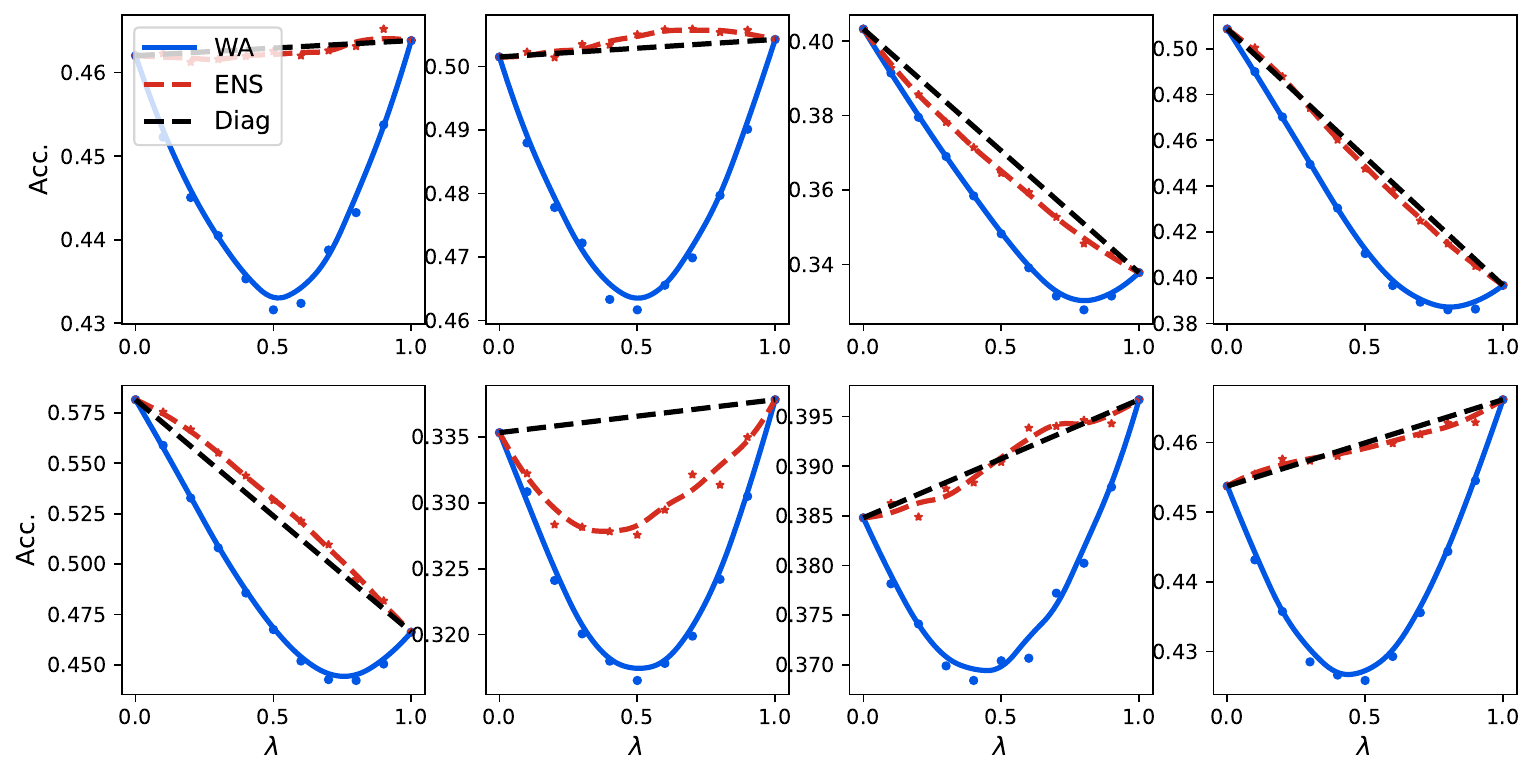}
		\vspace{-1em}
		\caption{Train (corrupt). More results enriching \Cref{fig:corrupt_wavsens_corrupttrain} with different pairs of RMs.}
		\label{fig:grid_wavsens_traincorrupt}%
		\includegraphics[width=0.55\textwidth]{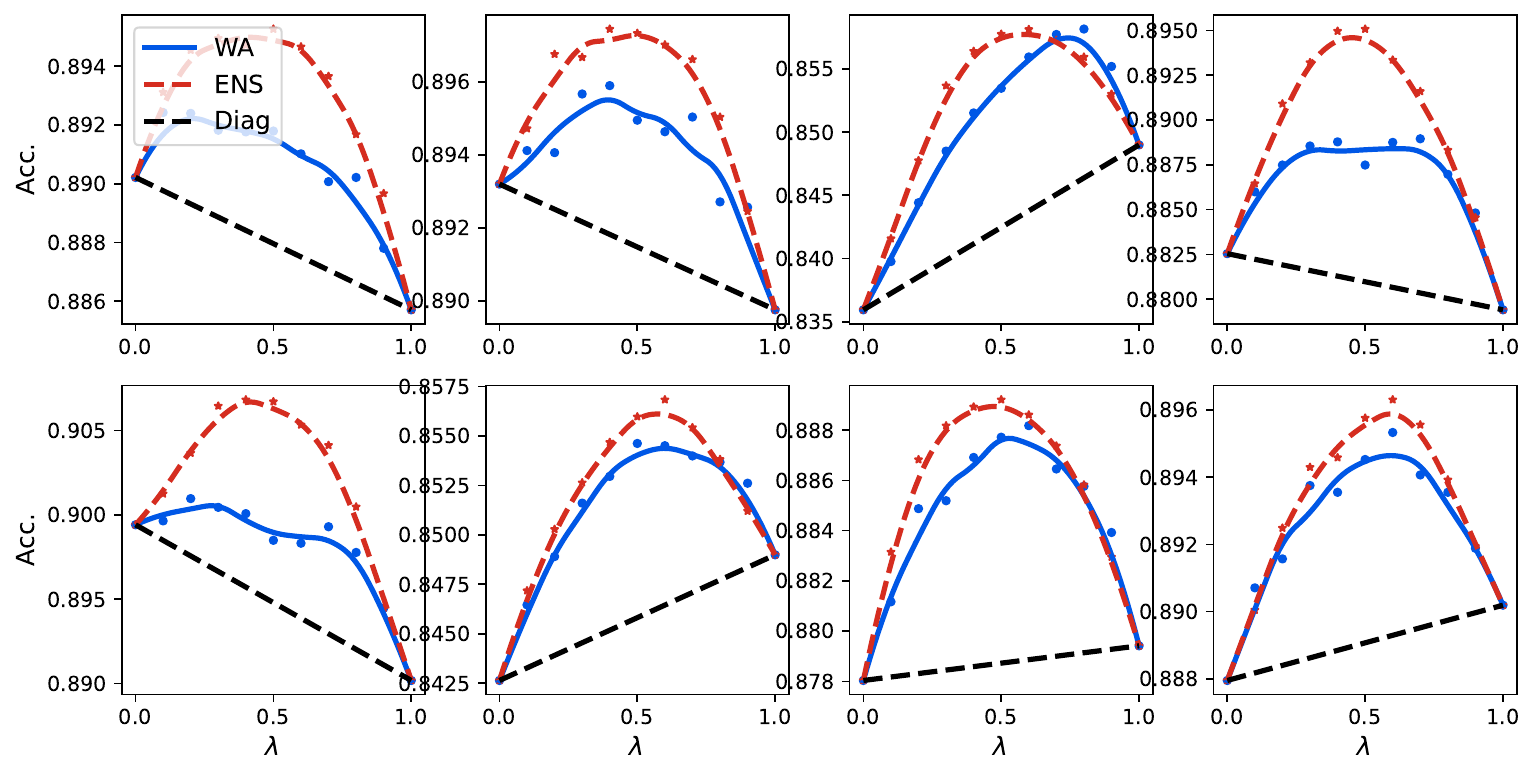}
		\vspace{-1em}
		\caption{Train (clean). More results enriching \Cref{fig:corrupt_wavsens_cleantrain} with different pairs of RMs.}
		\label{fig:grid_wavsens_trainclean}%
		\includegraphics[width=0.55\textwidth]{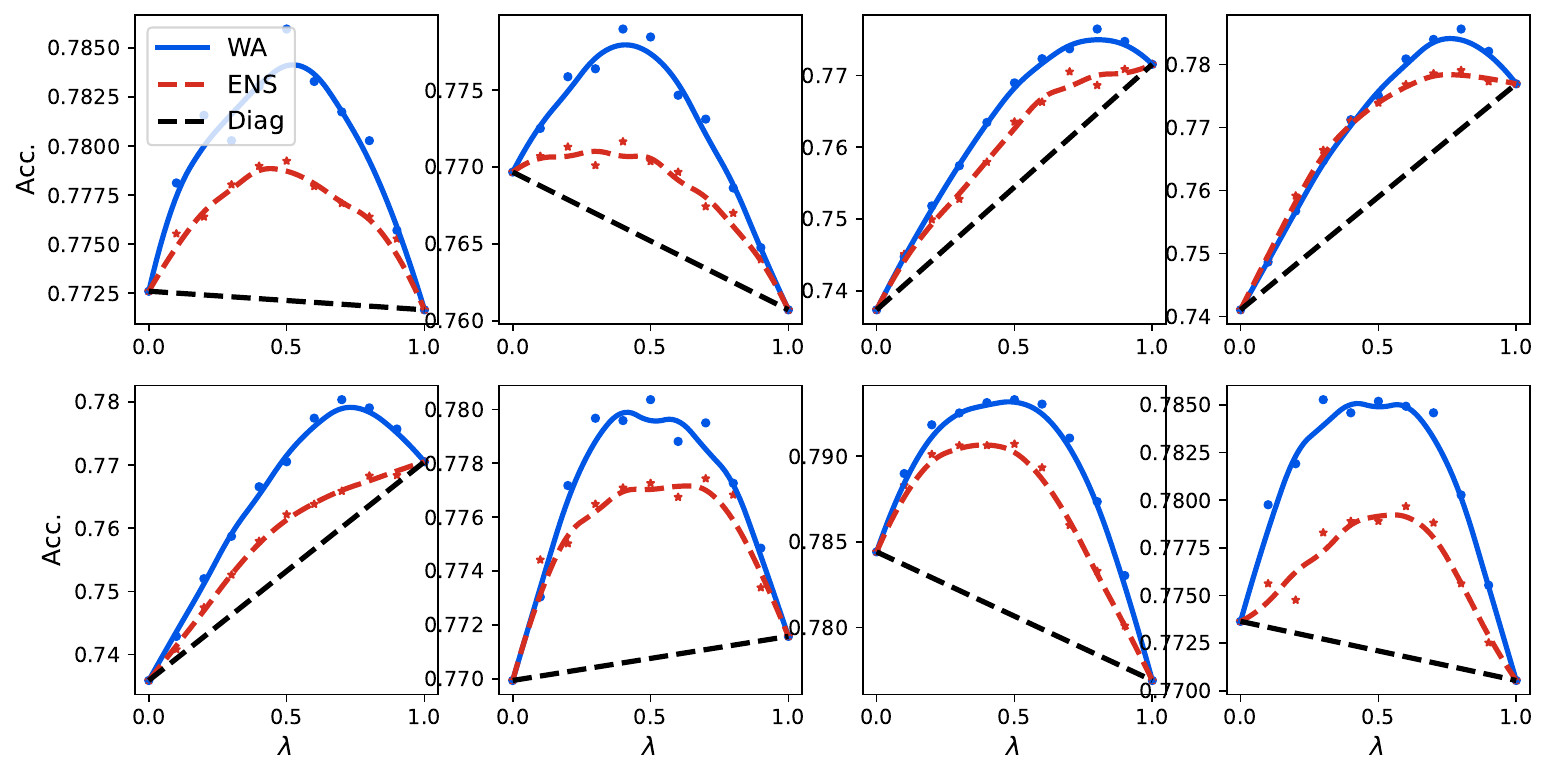}
		\vspace{-1em}
		\caption{Validation (ID). More results enriching \Cref{fig:corrupt_wavsens_valid} with different pairs of RMs.}
		\label{fig:grid_wavsens_valid}%
		\includegraphics[width=0.55\textwidth]{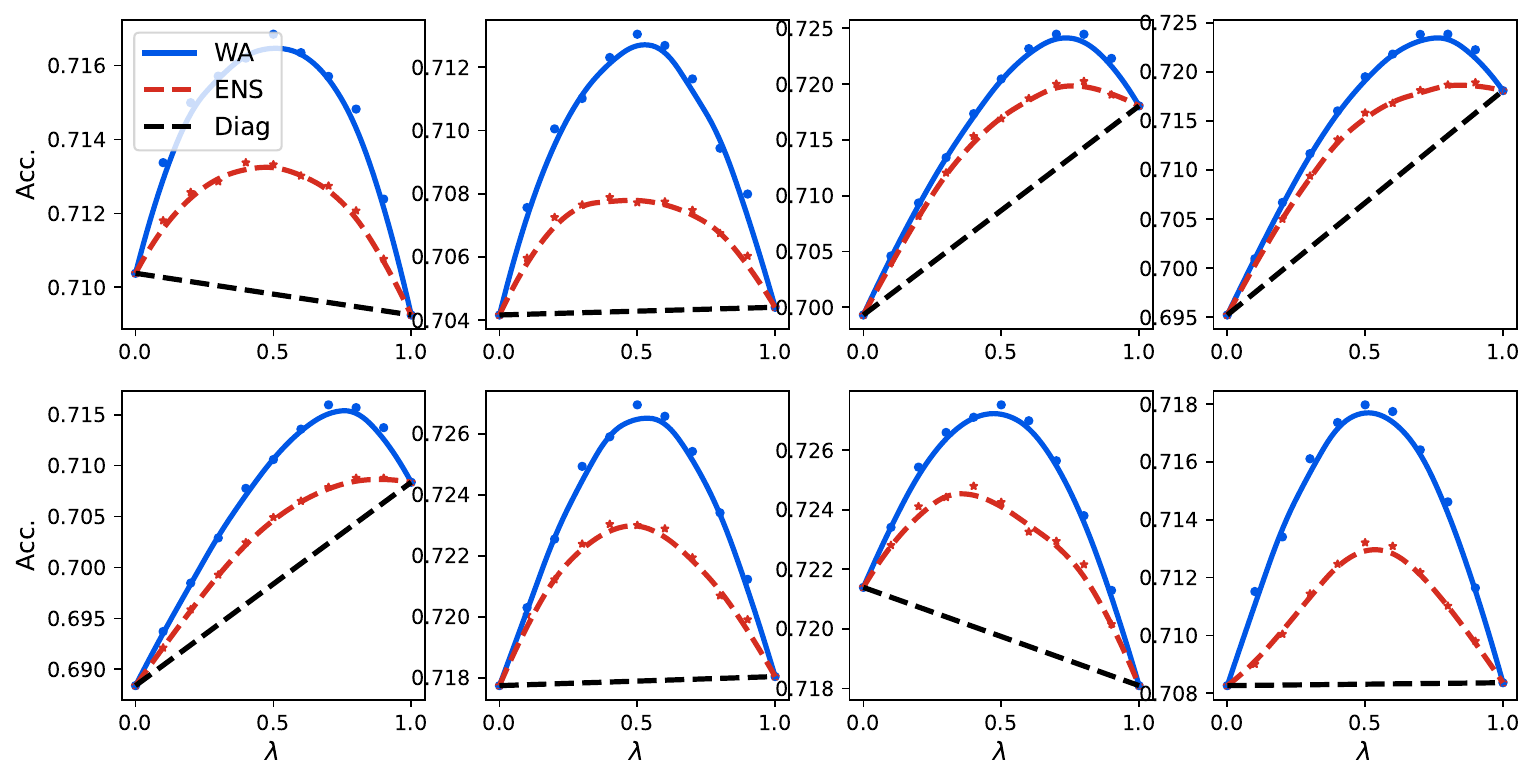}
		\vspace{-1em}
		\caption{Test (OOD). More results enriching \Cref{fig:corrupt_wavsens_ood} with different pairs of RMs.}
		\label{fig:grid_wavsens_testood}%
	\end{center}%
\end{figure}%
\clearpage
\FloatBarrier
\subsection{BoN experiments}
\begin{figure}[h!]
	\begin{center}
		\begin{subfigure}[b]{0.24\textwidth}
			\includegraphics[width=1.0\textwidth]{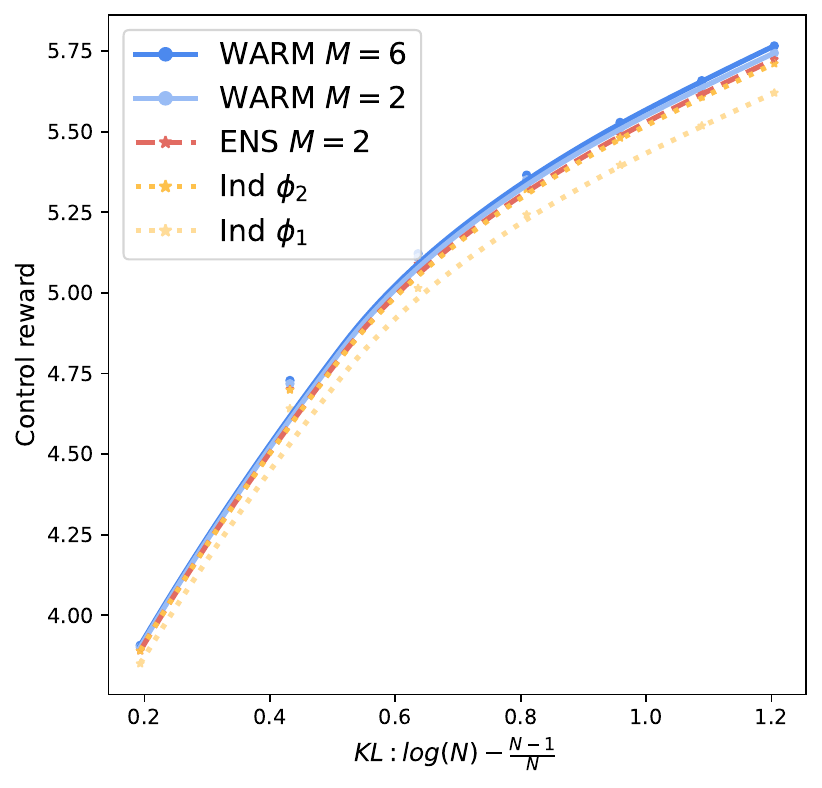}
			\caption{PaLM (clean).}
			\label{fig:bon8_cand_warm_score_selected_kl}%
		\end{subfigure}%
		\hfill
		\begin{subfigure}[b]{0.24\textwidth}
			\includegraphics[width=1.0\textwidth]{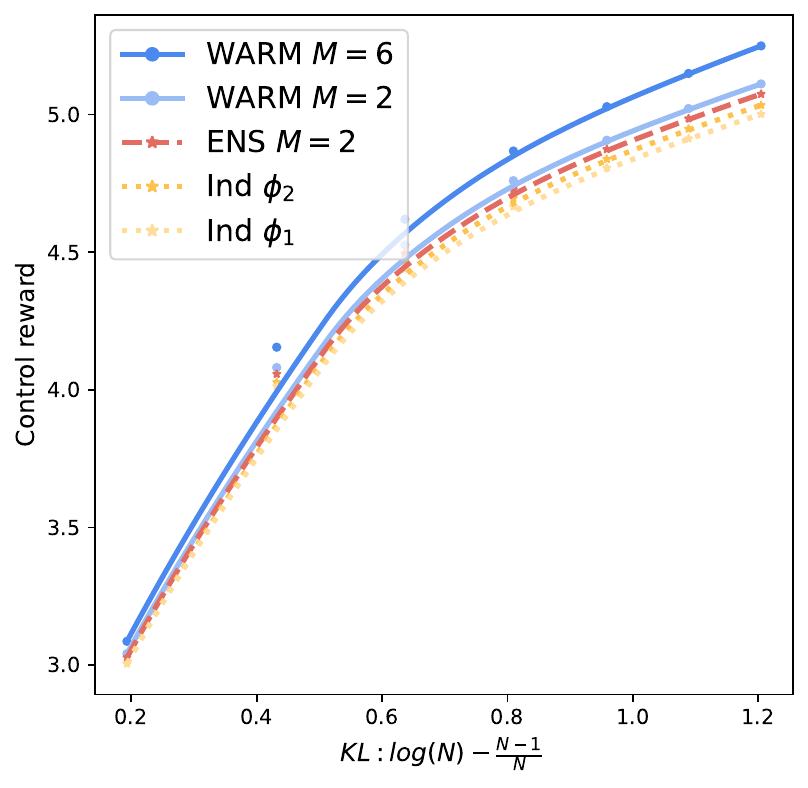}
			\caption{PaLM (corrupt).}
			\label{fig:bon8_cand_warmc_score_kl}%
		\end{subfigure}%
		\hfill
		\begin{subfigure}[b]{0.24\textwidth}
			\includegraphics[width=1.0\textwidth]{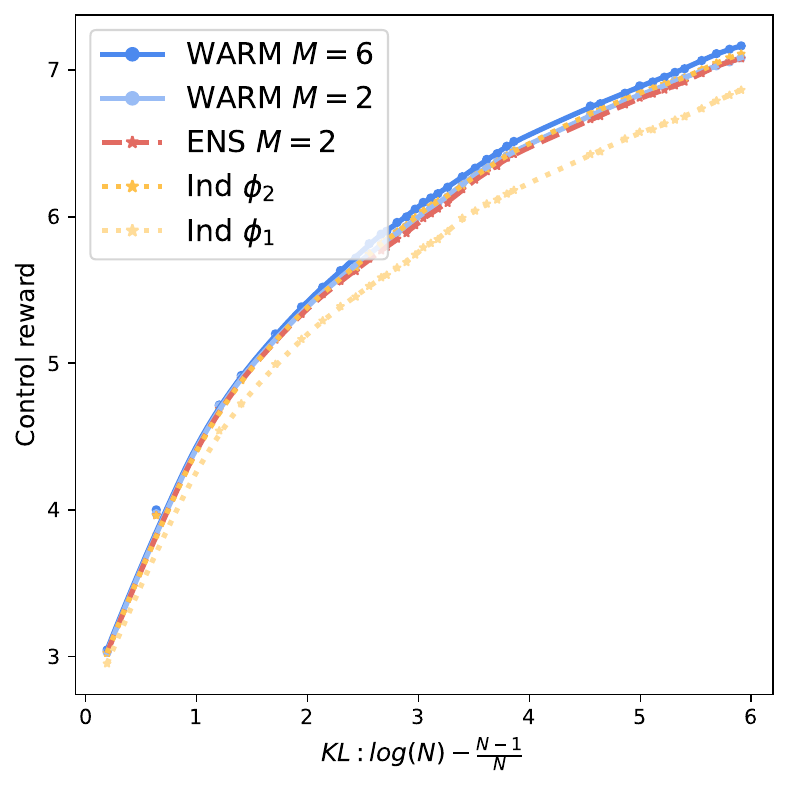}
			\caption{T5 (clean).}
			\label{fig:bon1000_t5x_warm_score_selected_kl}%
		\end{subfigure}%
		\hfill
		\begin{subfigure}[b]{0.24\textwidth}
			\includegraphics[width=1.0\textwidth]{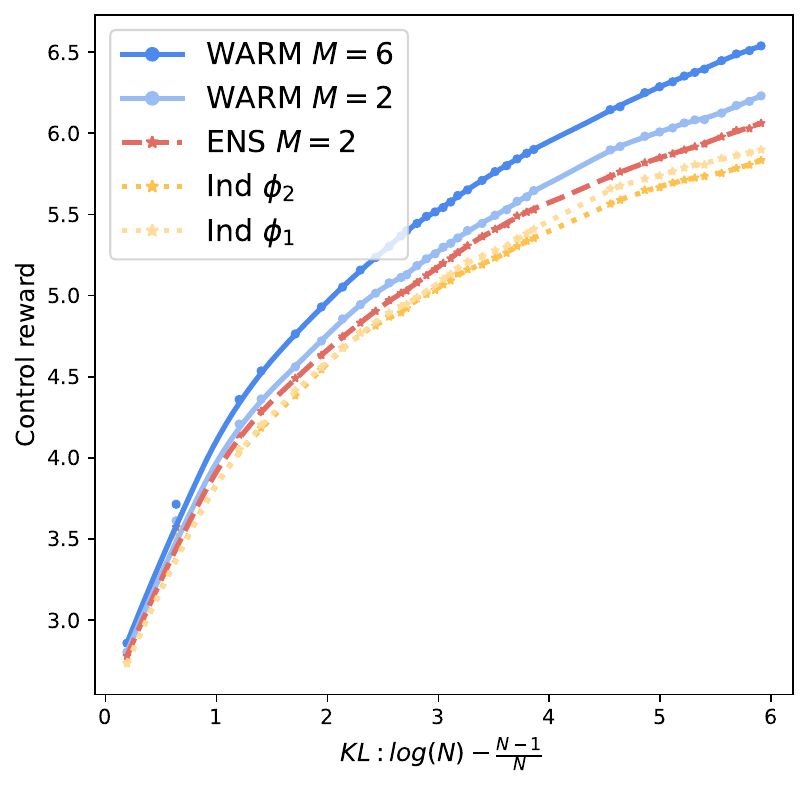}
			\caption{T5 (corrupt).}
			\label{fig:bon1000_t5x_warmc_score_kl}%
		\end{subfigure}%
	\end{center}%
	\caption{Same as \Cref{fig:controlvskl_hacking}, but with \textbf{absolute values of the control reward for BoN experiments}. We consider two SFT policies to generate candidate summaries: one based on PaLM architecture \cite{anil2023palm}, the other on T5 architecture \cite{JMLR:v21:20-074}. In both cases, we observe that \WARM performs better than ENS and the individual networks in terms of pointwise control RM.}%
	\label{fig:bon_abs}%
\end{figure}%
\begin{figure}[h!]
	\begin{center}
		\begin{subfigure}[b]{0.325\textwidth}
			\includegraphics[width=1.0\textwidth]{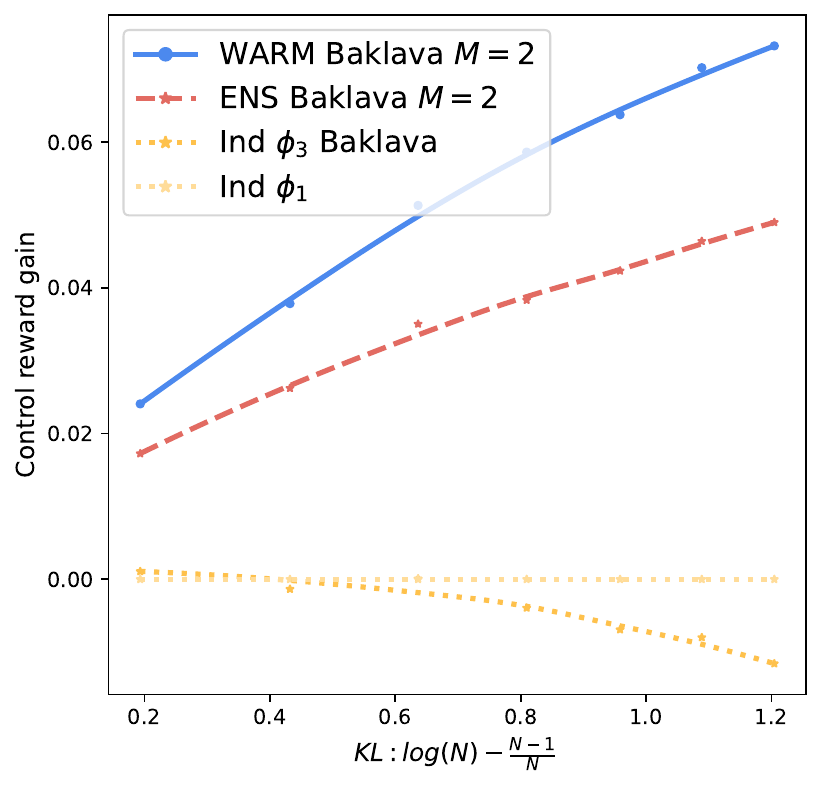}
			\caption{\BAK with PaLM.}
			\label{fig:bon8_cand_warm_scoregain_baklava_kl}%
		\end{subfigure}%
            \hspace{2.25cm}
		\begin{subfigure}[b]{0.325\textwidth}
			\includegraphics[width=1.0\textwidth]{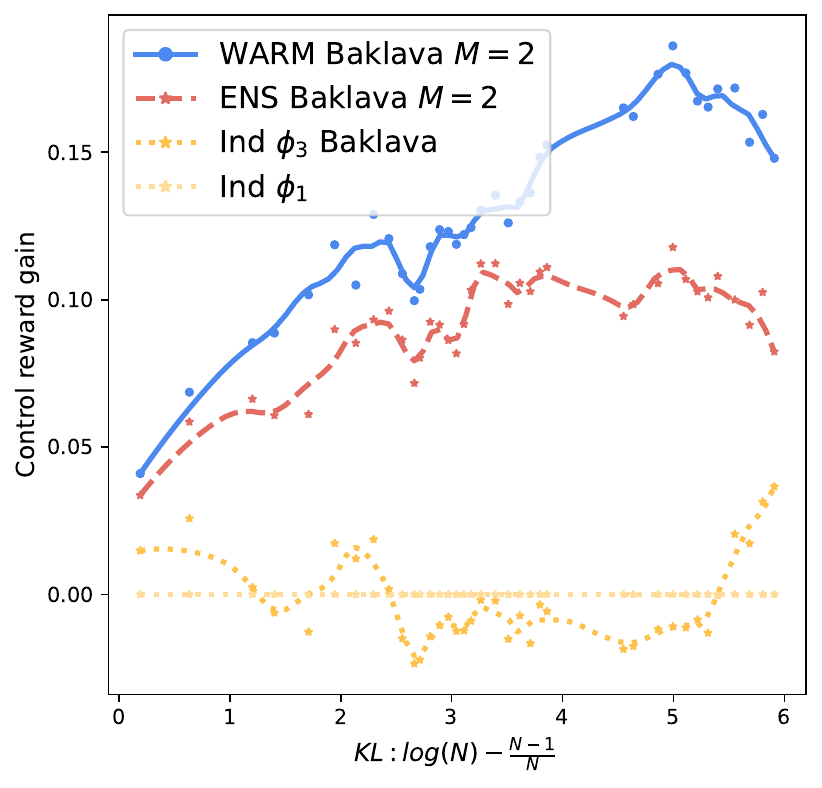}
			\caption{\BAK with T5.}
			\label{fig:bon1000_t5x_warm_scoregain_baklava_kl}%
		\end{subfigure}%
	\end{center}%
	\caption{\textbf{Control reward for BoN experiments} (clean setup) with \BAK when the two fine-tunings $\phi_1$ and $\phi_3$ have different featurizer initializations, collected respectively at steps 12k and 8k from a shared SFT.}%
	\label{fig:bon1000bak}%
\end{figure}%

\begin{figure}[h!]
	\begin{center}
		\begin{subfigure}[b]{0.325\textwidth}
			\includegraphics[width=1\textwidth]{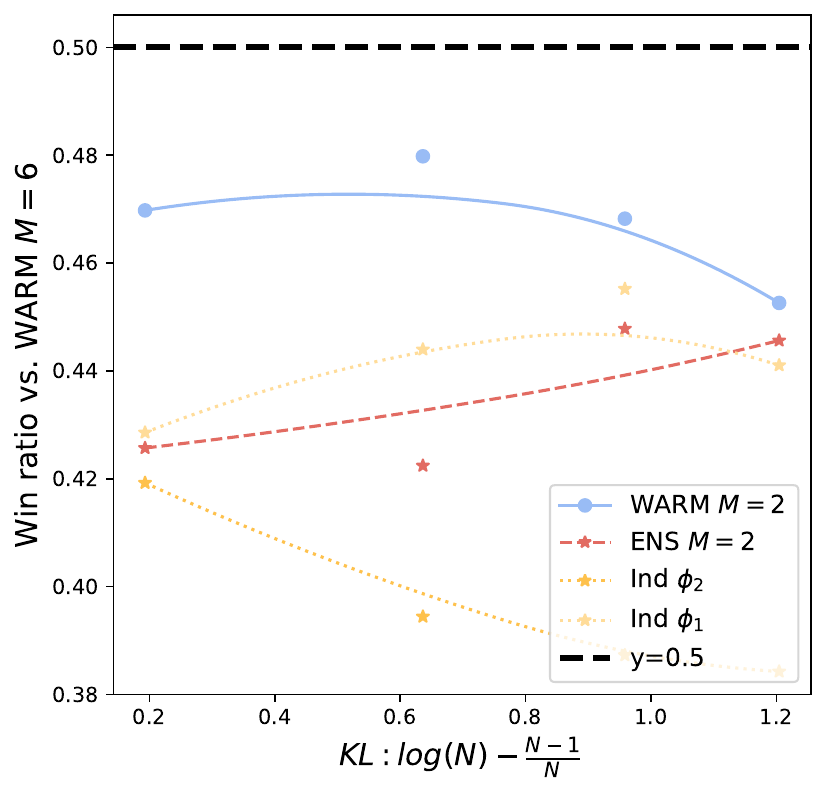}
			\caption{PaLM.}
			\label{fig:aipref_bon_cand_warm6}%
		\end{subfigure}%
            \hspace{2.25cm}%
		\begin{subfigure}[b]{0.325\textwidth}
			\includegraphics[width=1\textwidth]{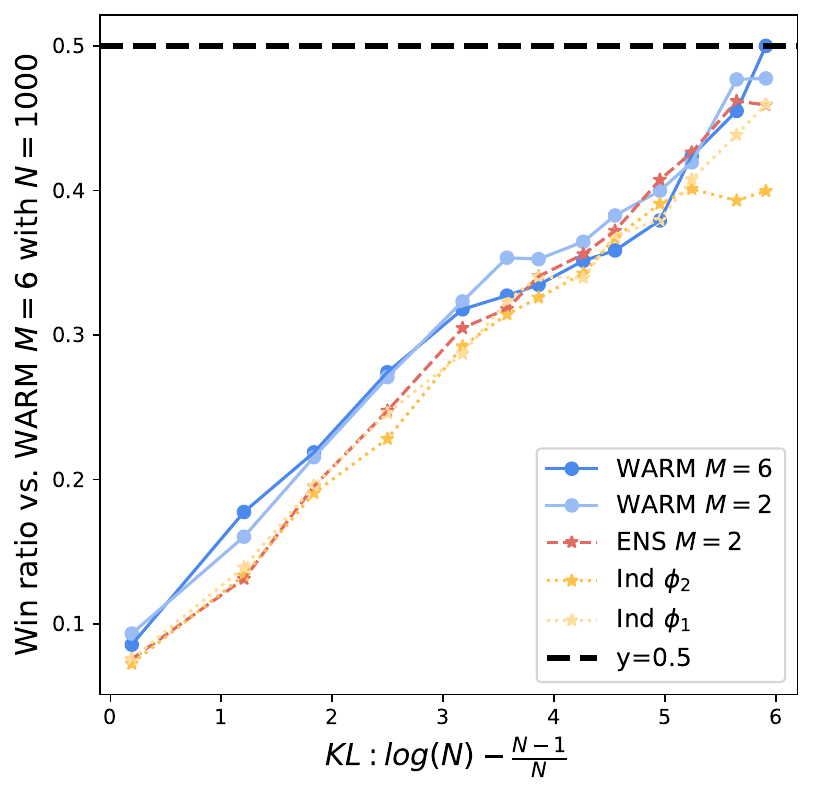}
			\caption{T5 \versus \WARM w/ $N=1000$.}
			\label{fig:bon_t5x_vswarm6n1000_kl}%
		\end{subfigure}%
	\end{center}
	\caption{\textbf{Oracle preference metric for BoN experiments} (clean setup).
		\Cref{fig:aipref_bon_cand_warm6} confirms \Cref{fig:aipref_bon_t5x_warm6} but on generations from PaLM SFT.
		\Cref{fig:bon_t5x_vswarm6n1000_kl} shows win rates for BoN on T5 generations for \WARM with $M=6$ and always $N=1000$ for BoN \versus other RMs with $1\leq N\leq 1000$. We validate that BoN limits reward hacking compared to RL, as performances get better when increasing $N$.}
	\label{fig:bon_aipref_ablation}%
\end{figure}%
\FloatBarrier
\subsection{RL experiments}
\FloatBarrier
\subsubsection{Experiments with corrupted preference dataset}
\FloatBarrier

\begin{figure}[h!]
	\begin{center}
		\includegraphics[width=0.5\textwidth]{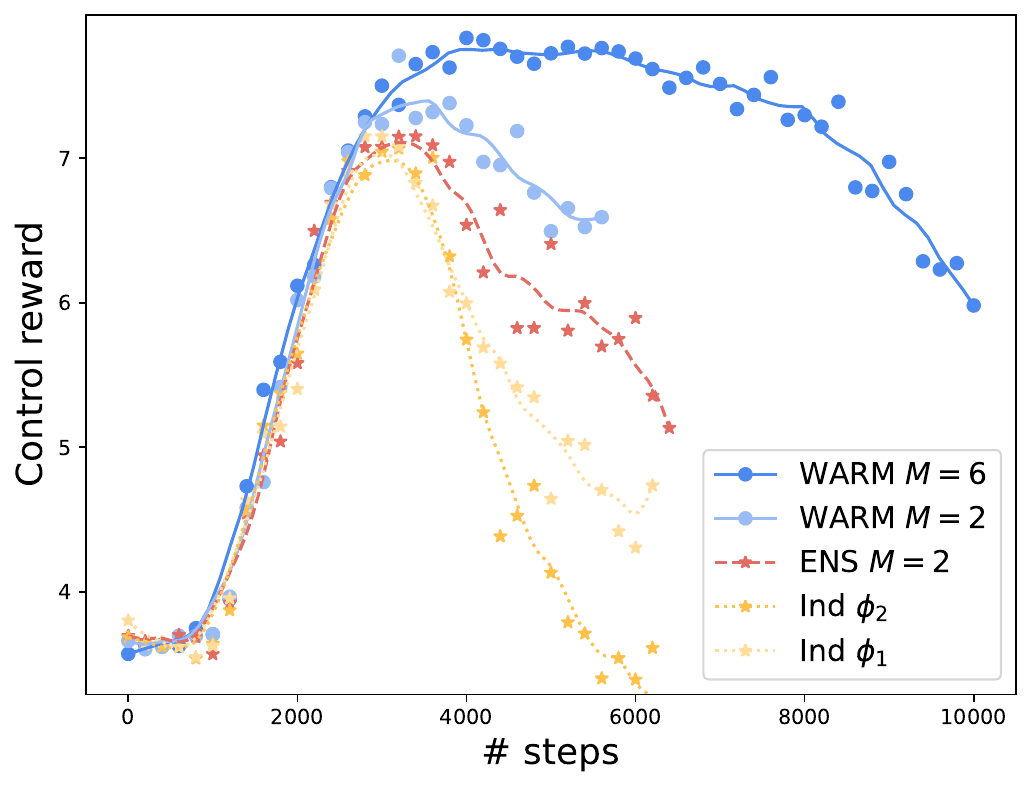}
	\end{center}%
	\caption{\textbf{RL experiments}. Same as \Cref{fig:main:hacking} but with 25\% corruption in the preference dataset.}%
	\label{fig:controlvsstep_corruptalp}%
\end{figure}%
\FloatBarrier
\newpage
\subsubsection{Experiments with clean preference dataset}
\begin{figure}[h!]
	\begin{center}
		\begin{subfigure}[b]{0.45\textwidth}
			\includegraphics[width=1\textwidth]{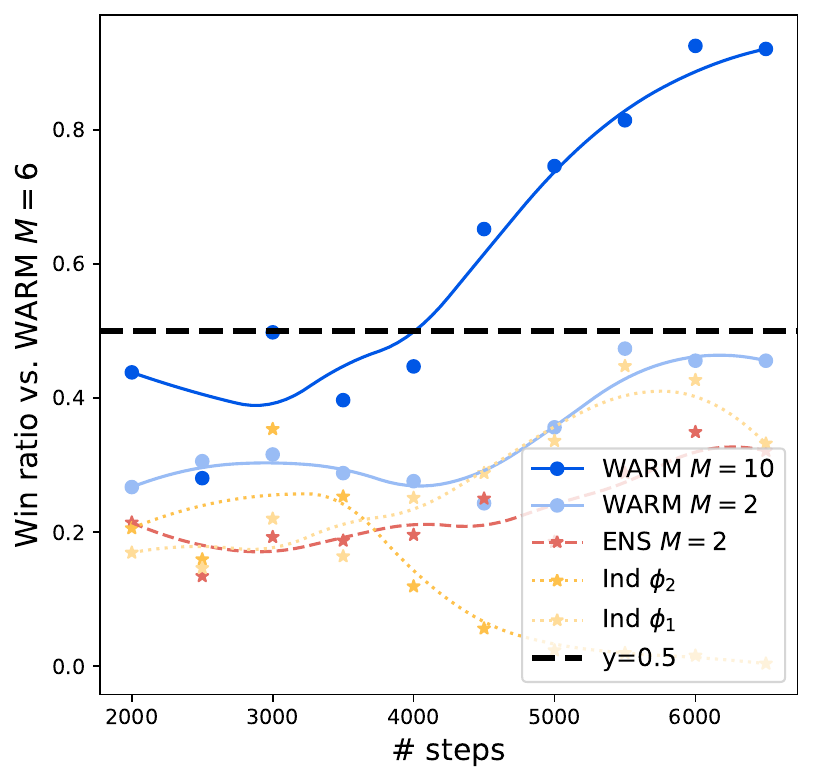}
			\caption{\WARM $M=6$.}
			\label{fig:aipref_warm6_sametrainingsteps}%
		\end{subfigure}%
		\hfill%
		\begin{subfigure}[b]{0.45\textwidth}
			\includegraphics[width=1\textwidth]{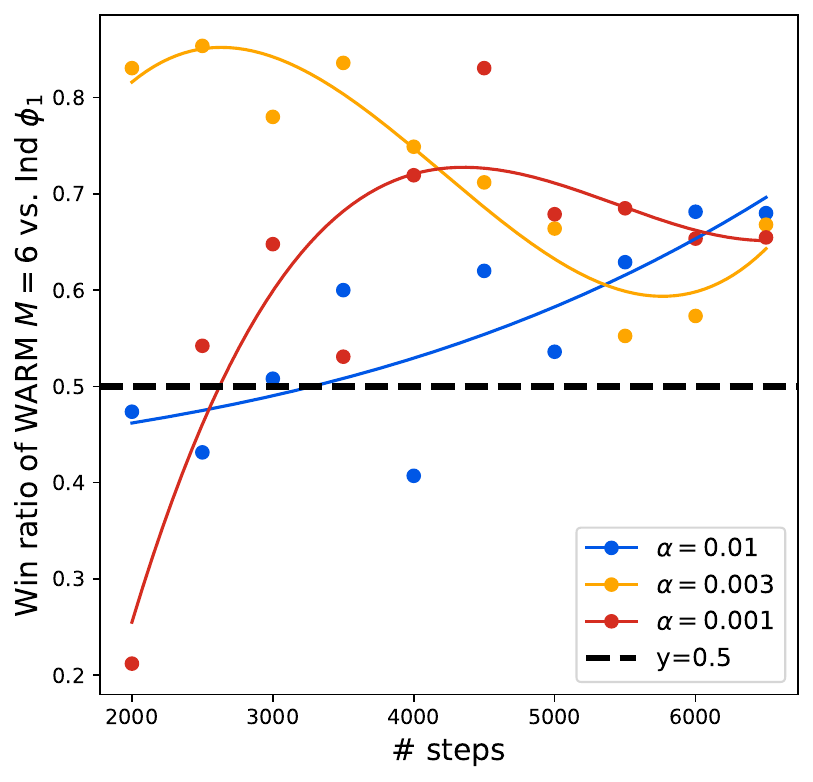}
			\caption{Impact of $\alpha$.}
			\label{fig:aipref_warm6vsbest_alphas}%
		\end{subfigure}%
	\end{center}%
	\caption{\textbf{Oracle preference metric for RL experiments} at fixed number of training steps (clean setup). \Cref{fig:aipref_warm6_sametrainingsteps} plots the win rate of the policy with \WARM $M=6$ \versus the other policies, all at the same number of training steps. \Cref{fig:aipref_warm6vsbest_alphas} shows the win rate of \WARM $M=6$ against the policy trained with a single RM $\phi_1$ (the best according to OOD accuracy) along training for different values of $\alpha$ controlling the \KL regularization strength.}%
	\label{fig:rl_aipref_fixedepochs}%
\end{figure}%
\begin{figure}[h!]
	\begin{center}
		\begin{subfigure}[b]{0.40\textwidth}
			\includegraphics[width=1.0\textwidth]{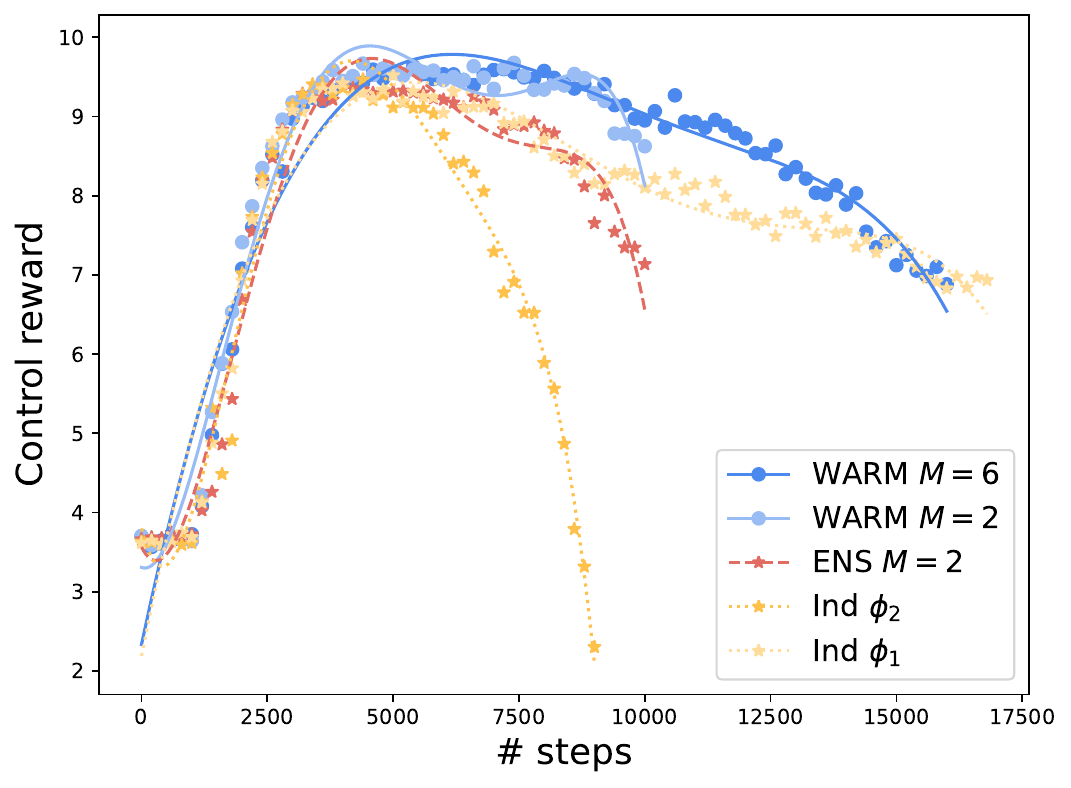}
			\caption{Control reward \versus training steps.}
			\label{fig:controlvsstep_ALP01}%
		\end{subfigure}%
            \hspace{1cm}
		\begin{subfigure}[b]{0.40\textwidth}
			\includegraphics[width=1.0\textwidth]{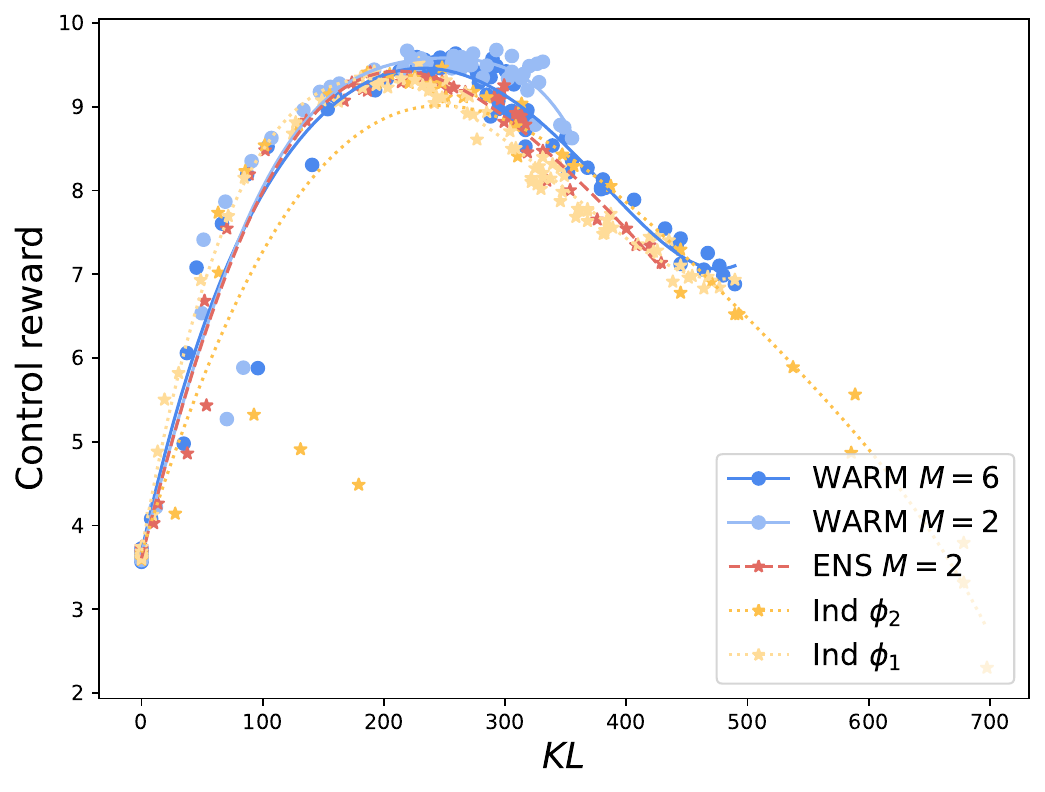}
			\caption{Control reward \versus \KL.}
			\label{fig:controlvskl_ALP01}%
		\end{subfigure}%
	\end{center}%
	\caption{\textbf{Control reward for RL experiments} with $\alpha=0.01$ (clean setup).}%
	\label{fig:rl01}%
\end{figure}%
\FloatBarrier
\begin{figure}[h!]
	\begin{center}
		\begin{subfigure}[b]{0.40\textwidth}
			\includegraphics[width=1.0\textwidth]{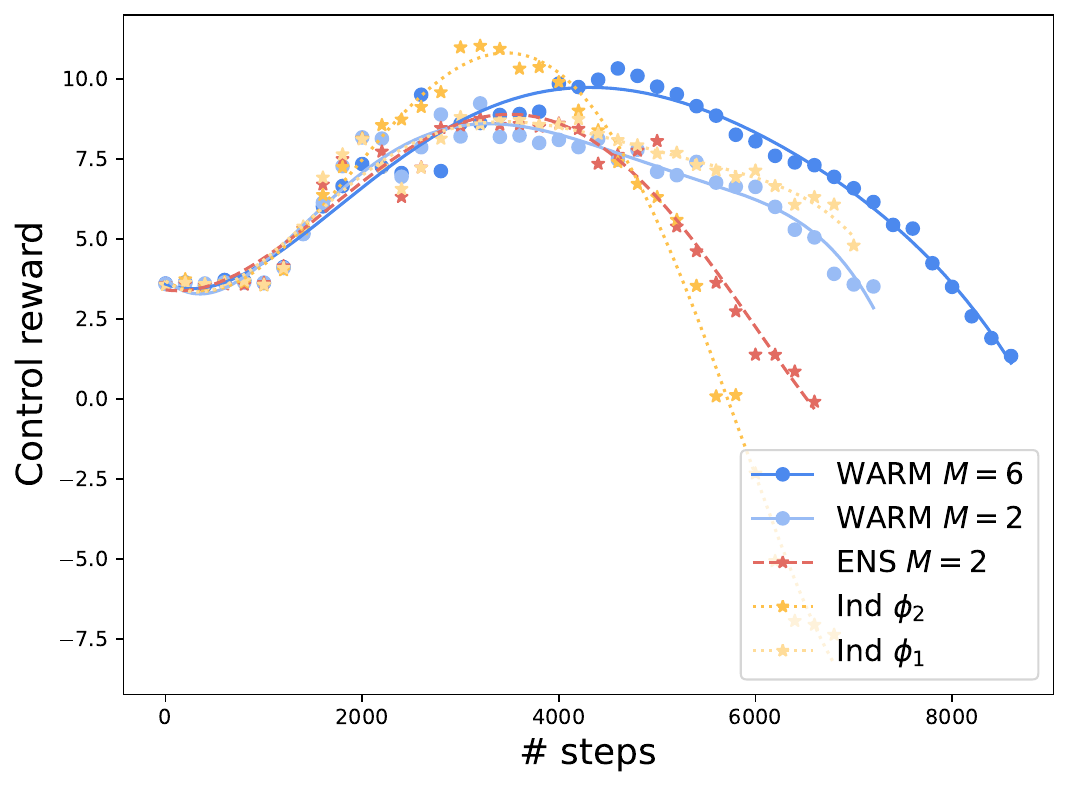}
			\caption{Control reward \versus training steps.}
			\label{fig:controlvsstep_ALP001}%
		\end{subfigure}%
            \hspace{1cm}
		\begin{subfigure}[b]{0.40\textwidth}
			\includegraphics[width=1.0\textwidth]{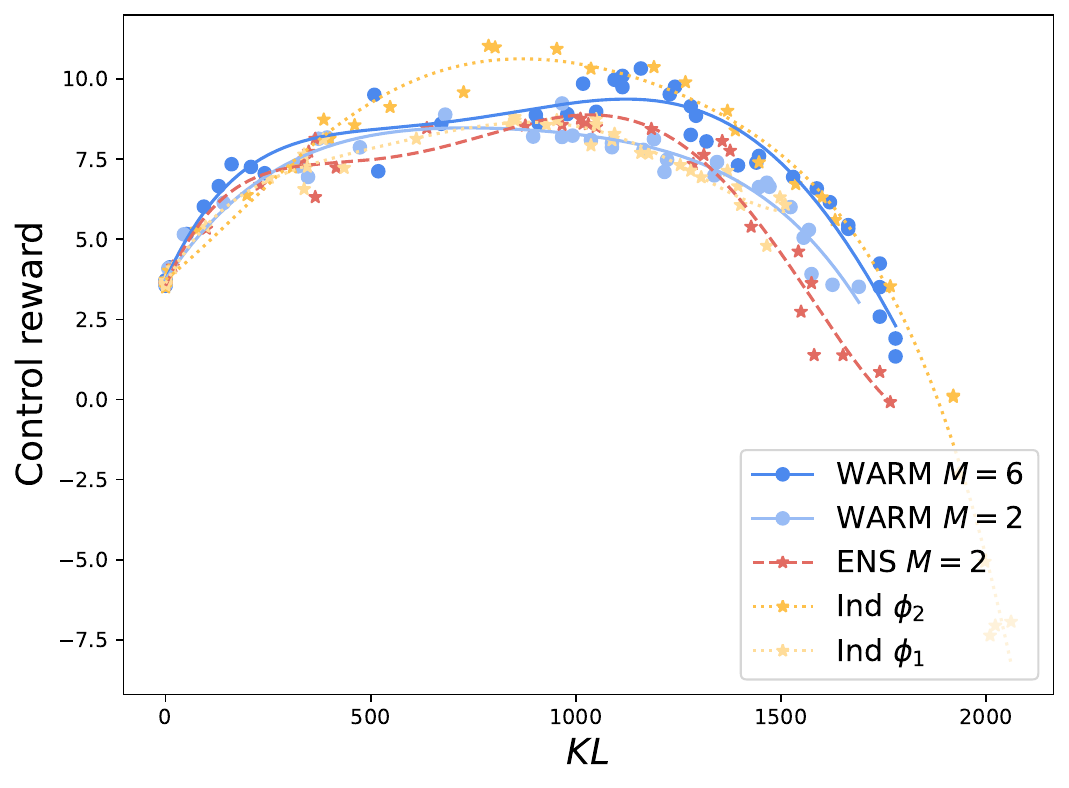}
			\caption{Control reward \versus \KL.}
			\label{fig:controlvskl_ALP001}%
		\end{subfigure}%
	\end{center}%
	\caption{\textbf{Control reward for RL experiments} with $\alpha=0.001$ (clean setup).}%
	\label{fig:rl001}%
\end{figure}%
\FloatBarrier
\subsection{Distillation experiments}
\label{app:distillation}
\begin{figure}[h!]
	\begin{center}
		\includegraphics[width=0.5\textwidth]{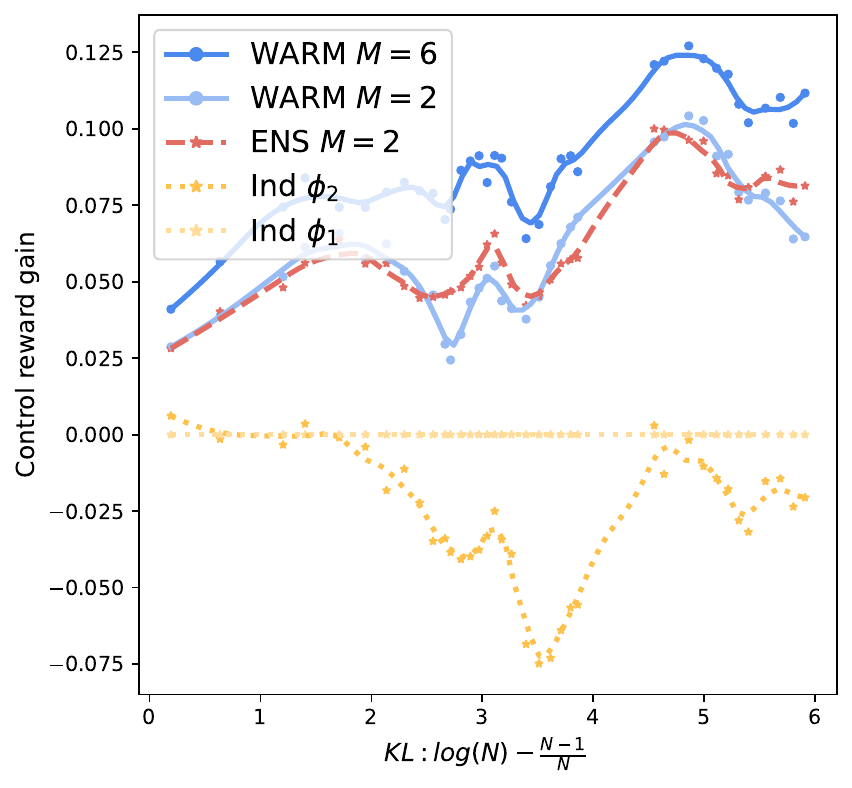}
		\caption{\textbf{BoN experiment in the distillation setup from \cite{gao2022scaling}}.
		The labels in the preference dataset are given by the control RM, the same RM which gives the $y$-axis.
		The candidate summaries are generated by a SFT with the T5 architecture \cite{JMLR:v21:20-074}.
		The blue lines represent \WARM with $M$ weights: \WARM performs higher than the individual RMs (in yellows) or when ensembling their predictions (ENS in red).}
		\label{fig:distill}
	\end{center}
\end{figure}

In \Cref{fig:distill} we reproduce the distillation setup from \cite{gao2022scaling}, where the control PaLM-XS RM generates the labels to train PaLM-XXS RMs.
As a side note, we observed that distillation changes the diversity across fine-tuned RMs, thus potentially altering the significance of the distillation setup, motivating us in exploring the more realistic RLAIF setup.

\end{document}